\documentclass[a4paper,fleqn]{cas-dc}
\usepackage{tikz}
\usetikzlibrary{arrows,fit}
\usetikzlibrary{mindmap,trees,shapes.geometric}
\usepackage{longtable}
\usepackage{hyperref}
\usepackage{orcidlink}
\usetikzlibrary{positioning, arrows.meta, shapes, backgrounds}
\usepackage[x11names]{xcolor}
\usepackage[colorinlistoftodos,prependcaption,textsize=small]{todonotes}
\usepackage[most]{tcolorbox}
\usepackage{tabularx}
\usepackage{booktabs}
\usepackage{threeparttable}
\usepackage{tikz}
\usetikzlibrary{arrows.meta, positioning, shapes.geometric}
\usetikzlibrary{shadows,trees}
\usetikzlibrary{calc}

\usepackage{forest}
\forestset{
  default preamble={
    for tree={
      font=\sffamily,
      edge={-stealth},
      align=center,
      anchor=west,
    }
  }
}

\usepackage{array}
\usepackage{varwidth}

\definecolor{MyBlue}{RGB}{26,13,171}

\usepackage[authoryear]{natbib}

\usepackage{tikz-qtree}
\usepackage{adjustbox}

\usepackage{caption}

\def\tsc#1{\csdef{#1}{\textsc{\lowercase{#1}}\xspace}}
\tsc{}
\tsc{}

\begin{document}
\let\WriteBookmarks\relax
\def\floatpagepagefraction{1}
\def\textpagefraction{.001}

\shorttitle{}    

\shortauthors{}  

\title [mode = title]{Generalization vs. Specialization: Evaluating Segment Anything Model (SAM3) Zero-Shot Segmentation Against Fine-Tuned YOLO Detectors}


\tnotetext[2]{The publication of the article in OA mode was financially supported by ...} 

%


\author[1]{Ranjan Sapkota}[orcid=0000-0002-5417-6744]
\cormark[1]
\ead{rs2672@cornell.edu}
\affiliation[1]{organization={Cornell University, Department of Biological and Environmental Engineering},city={Ithaca},postcode={14850},state={NY},country={USA}}

\author[2]{Konstantinos I. Roumeliotis}[orcid=0000-0002-8098-1616]
\cormark[1]
\ead{k.roumeliotis@uop.gr}
\affiliation[2]{organization={University of the Peloponnese, Department of Informatics and Telecommunications},city={Tripoli},postcode={22131},country={Greece}}

\author[1]{Manoj Karkee}[orcid=0000-0001-5337-4848]
\ead{mk2684@cornell.edu}
\cormark[1]

\author[2]{Nikolaos D. Tselikas}[orcid=0000-0001-5799-3558]
\ead{ntsel@uop.gr}
\cormark[1]

\cortext[1]{Corresponding authors}



\begin{abstract}
Deep learning has advanced two fundamentally different paradigms for instance segmentation: specialized models optimized through task-specific fine-tuning and generalist foundation models capable of zero-shot segmentation. This work presents a comprehensive comparison between Meta’s SAM3 (Segment Anything Model, also called SAMv3) operating in zero-shot mode and three variants of Ultralytics’ YOLO11 (nano, medium, and large) fine-tuned for instance segmentation. The evaluation is conducted on the MinneApple dataset, a dense benchmark comprising 670 orchard images with 28,179 annotated apple instances, enabling rigorous validation of model behavior under high object density and occlusion. Our study uncovers critical methodological insights, showing that commonly used IoU thresholds can introduce artificial performance gaps of up to 30\%. Through systematic threshold sensitivity analysis, we identify IoU = 0.15 as a more appropriate criterion for dense instance segmentation, ensuring fair and meaningful comparison across models. Under this setting, the fine-tuned YOLO11 variants achieve F1 scores of 68.9\% (nano), 72.2\% (medium), and 71.9\% (large), highlighting their strong specialization. In contrast, SAM3 attains 59.8\% F1 in pure zero-shot mode, demonstrating notable generalization despite no domain-specific training. Crucially, IoU sensitivity analysis reveals a previously unrecognized trade-off: while YOLO11 models exhibit steep performance degradation across IoU thresholds (48 - 50 point drops from 0.10 to 0.50), SAM3 shows remarkable stability with only a 4-point drop, indicating 12-fold superior boundary quality despite lower absolute detection rates. This finding suggests that foundation models produce spatially precise masks even when missing instances, whereas specialized models prioritize detection completeness over boundary accuracy. We provide open-source code, evaluation pipelines, and methodological recommendations, contributing to a deeper understanding of when specialized fine-tuned models or generalist foundation models are preferable for dense instance segmentation tasks. This project repository is available on GitHub (~\href{https://github.com/Applied-AI-Research-Lab/Segment-Anything-Model-SAM3-Zero-Shot-Segmentation-Against-Fine-Tuned-YOLO-Detectors}{\texttt{Link}}).
\end{abstract}





\maketitle
\scriptsize
\normalsize

\section{Introduction}
\label{sec:Introduction}
Modern computer vision has entered a phase where scientists, engineers, and practitioners must routinely choose between two very different design philosophies: highly specialized models fine-tuned for a single task, and large generalist foundation models intended to ``work everywhere'' \cite{awais2025foundation,liu2025toward}. Nowhere is this dilemma more evident than in dense instance segmentation problems \cite{zhang2025wheat3dgs, wang2025automated, liu2025simplemask}, where accurate delineation of many small, overlapping objects is required under complex, real-world conditions \cite{szeliski2022computer}. While numerous architectures and toolkits are now available, there remains a practical knowledge gap: researchers and practitioners often lack clear, data-driven guidance on \emph{when} to deploy specialized versus generalist models, \emph{what} capabilities to expect from each family, and \emph{how} to configure them for reliable use in real applications.

\begin{figure*}[h!]
    \centering
    \includegraphics[width=1.8\columnwidth]{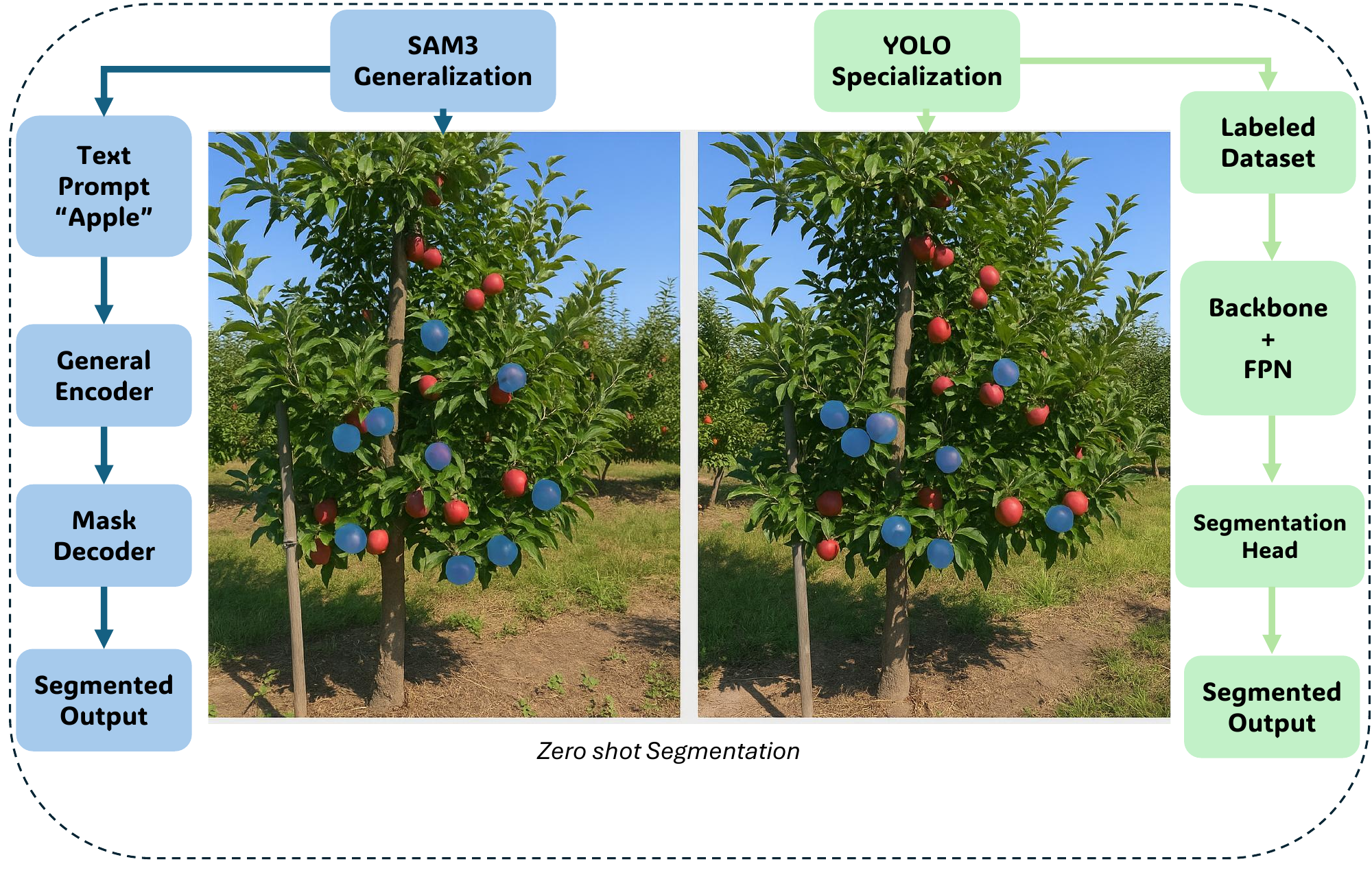}
    \caption{\small
    Conceptual overview: Side-by-side apple segmentation comparison illustrating SAM3 generalization (Prompt → Encoder → Decoder → Output) versus YOLO specialization (Labeled Data → Backbone + FPN → Segmentation Head → Output) under identical orchard conditions.
    }
    \label{fig:intro_overview}
\end{figure*}

Recent advances in deep learning have sharpened the distinction between two fundamentally different paradigms for instance segmentation, a contrast visually summarized in Figure~\ref{fig:intro_overview}. On one side are specialized, task-specific models that rely on supervised training with curated datasets and are explicitly optimized for a particular domain. This family includes both two-stage and one-stage architectures. Early two-stage detectors such as Faster R-CNN~\citep{ren2015faster} established the core region-proposal framework by first generating candidate object regions and then refining them through dedicated classification and regression heads. Mask R-CNN~\citep{he2017mask} extended this paradigm with a parallel segmentation branch that enabled pixel-accurate instance masks, becoming a widely adopted baseline. Successive variants including Cascade R-CNN~\citep{cai2018cascade}, Cascade Mask R-CNN~\citep{wu2020object, he2021recognition}, PANet~\citep{liu2018path}, and Hybrid Task Cascade (HTC)~\citep{chen2019hybrid} enhanced multi-scale feature fusion, hierarchical refinement, and robustness across complex visual scenes.

In parallel, one-stage detectors such as RetinaNet~\citep{lin2017focal} introduced focal loss to mitigate class imbalance, enabling fast, high-performance instance segmentation pipelines suitable for real-time applications. Models in this specialized category follow workflows analogous to the YOLO pipeline illustrated in Figure~\ref{fig:intro_overview}, which emphasizes supervised learning from labeled datasets, backbone-driven representation learning, and task-tuned segmentation heads. Contrasting with these specialization-driven approaches, emerging foundation models adopt a generalization-first philosophy in which segmentation is conditioned on high-level prompts rather than domain-specific training. This conceptual divergence generalization via prompt-based inference versus specialization via supervised optimization underscores the evolving landscape of instance segmentation and motivates rigorous comparative evaluations.

Building on these foundations, a rich ecosystem of real-time and near real-time instance segmentation models has emerged. YOLACT~\citep{bolya2019yolact} and YOLACT++~\citep{zhou2020yolact++} decouple prototype mask generation from per-instance mask coefficients, enabling fast, single-shot instance segmentation. Likewise, CenterMask~\citep{lee2020centermask} combines the ideas of anchor-free center-based detection with strong mask heads, while FCIS and subsequent work explored fully convolutional instance segmentation without explicit region proposal networks. More recently, kernel-based and dynamic-filter approaches such as CondInst~\citep{tian2020conditional} and SOLO / SOLOv2~\citep{wang2020solov2, wang2020solo} formulated instance segmentation as dense category-aware mask prediction at grid locations, eliminating the need for bounding box proposals altogether. In parallel, transformer-based methods such as DETR~\citep{carion2020end} and segmentation-oriented successors like Mask2Former~\citep{cheng2022masked} treat detection and segmentation as a set prediction problem, offering a unified framework that can be specialized through fine-tuning on domain-specific data.

Within this broader landscape, the YOLO (You Only Look Once) family, originally proposed for real-time object detection~\citep{redmon2016you}, has evolved into a suite of architectures that support both detection and instance segmentation. Successive generations (e.g., YOLOv3–v8, YOLOv9, YOLOv10) and derivatives such as YOLOv5-seg, YOLOv7-mask, and YOLOv8-seg incorporate multi-scale feature pyramids, decoupled heads, and improved segmentation branches while maintaining high throughput \cite{sapkota2025yolo, sapkota2025ultralytics}. The latest generation, YOLO11~\citep{yolo11_2024}, continues this trajectory with enhanced feature pyramids, anchor-free detection heads, and efficient backbones, and is released in multiple capacity tiers (e.g., nano, medium, large) to target different accuracy-efficiency trade-offs. When fine-tuned on a specific dataset, these specialized models can exploit domain-specific regularities in object appearance (e.g., apple shape and texture in orchard images), background structure (foliage, branches, sky), and data distribution (viewpoints, scales), often achieving state-of-the-art performance for that particular task while remaining computationally efficient for real-time deployment~\citep{yolo11_2024}.

On the other side are generalist foundation models~\citep{bommasani2021opportunities} trained at scale to support zero-shot or few-shot inference across many tasks without additional supervised training. In computer vision, the Segment Anything Model (SAM)~\citep{kirillov2023segment, carion2025sam} was a landmark: trained on SA-1B (over 1 billion masks across 11 million images), SAM can segment arbitrary objects in new images when prompted with points, boxes, or text. The latest iteration, SAM3~\citep{carion2025sam}, refines this paradigm with more efficient transformer backbones, enhanced prompt encoders, and improved mask decoders, further strengthening its zero-shot capabilities \cite{sapkota2025the}. From a practitioner’s perspective, such models are attractive because they eliminate the need for costly annotation and training pipelines, and they promise robustness across diverse domains with minimal configuration.

This emergence of increasingly powerful segmentation frameworks has brought forward a practical and still unresolved question for the vision community: given a realistic, densely populated segmentation task and a publicly available benchmark, should practitioners invest in collecting labels and fine-tuning a specialized model such as YOLO11, or rely on the zero-shot capabilities of a foundation model such as SAM3? Although numerous agricultural computer vision studies have shown that supervised, task-specific models can excel in fruit detection and segmentation~\citep{bargoti2017deep, koirala2019deep, tian2019apple}, and several surveys document the rapid adoption of deep learning across similar domains~\citep{kamilaris2018deep, liakos2018machine, gao2020deep}, these studies typically evaluate models in isolation. They do not provide clear, comparative evidence about when specialization genuinely outperforms generalization or vice versa under controlled conditions. At the same time, the success of foundation models has raised the possibility that, in certain environments, the cost of annotation, training, and hyperparameter tuning may no longer be necessary, especially when zero-shot or prompt-based inference already yields competitive results.

\subsection{Research Gap and Problem Statement}
Despite the availability of modern tools such as YOLO11~\citep{yolo11_2024}, its recent other versions \cite{sapkota2025yolo}and SAM3~\citep{carion2025sam}, researchers and practitioners lack systematic, evidence-based guidance for three essential decisions.
\begin{itemize}
    \item Model selection: When is task-specific fine-tuning worth the engineering cost, and when is zero-shot segmentation sufficient for practical use?
    \item Performance interpretation: How do the strengths and limitations of specialized versus generalist models differ across both instance-level metrics (e.g., precision, recall, F1) and pixel-level fidelity measures?
    \item Evaluation protocol design: How should IoU thresholds and matching criteria be chosen so that evaluation reflects the realities of dense, cluttered natural-scene environments rather than artifacts of overly strict or misaligned metrics?
\end{itemize}

These questions remain largely unexplored in the literature because few studies compare specialized and foundation-model approaches under identical, fair, and reproducible conditions. This paper addresses that gap by designing a controlled experimental study on a publicly available dataset that is both realistic and broadly representative of many natural-scene segmentation challenges, enabling actionable insights that extend well beyond agricultural imagery.

Within this broader context, agricultural vision datasets provide an especially compelling testbed for studying \textit{generalization} versus \textit{specialization} trade-offs in segmentation models \cite{angarano2024domain, joshi2023standardizing, chiu2020agriculture}. Dense plant canopies with object occlusions \cite{zhao2022stress, zhang2024algorithm, itakura2020automatic}, fine-scale structures (leaves, fruits, branches) \cite{zhang2025leaf, chen2025robust, tian2024cabbagenet, huang2024segment}, and strong daily and seasonal variabilities (both environmental and biological) create exactly the kind of complex, non-stationary distributions where it is unclear a priori whether a carefully fine-tuned model (e.g., YOLO-style detectors) or a large, pre-trained foundation model (e.g., SAM2/SAM3) will be more reliable. From a systems perspective, agricultural deployments are also representative of many real-world settings in which data collection is complex and time-consuming, annotation budgets are limited, edge compute is constrained, and robustness to changing environmental conditions is critical \cite{zhang2020overview, pintus2025emerging, kalyani2024application}. However, practitioners currently lack principled, quantitative guidance on the most cost-effective strategy: whether to invest in collecting a huge amount of labeled data and fine tuning a specialized architecture or to rely on the zero-shot capabilities of a segment-anything-style model, possibly with minimal prompt engineering. This uncertainty is amplified in dense instance segmentation problems, where evaluation itself is non-trivial, and naive choices of metrics or thresholds can obscure the true behavior of competing approaches. Interestingly, our systematic IoU sensitivity analysis reveals that SAM3 exhibits remarkable boundary quality stability across thresholds (only 4-point F1 drop from IoU=0.10 to 0.50), whereas YOLO11 models show steep degradation (48-50 point drops), suggesting distinct trade-offs between absolute accuracy and mask quality. 

Public orchard benchmarks such as the MinneApple dataset~\citep{hani2020minneapple} make it possible to investigate these issues in a controlled and reproducible way. MinneApple contains 670 high-resolution orchard images with pixel-wise instance masks for 28{,}179 apples (roughly 40–45 instances per image), capturing dense layouts, heavy occlusion, variable illumination, and substantial appearance variation within and across scenes. In this work, we treat MinneApple not merely as an agricultural case study, but as a canonical example of a dense, cluttered, natural-image environment where both specialized YOLO-style detectors and generalist SAM-style foundation models are plausible contenders. By rigorously comparing these families on such data, we aim to derive methodological and practical insights that generalize to any domain with similar visual and operational characteristics, not only to orchard perception or agricultural vision.

Using the MinneApple dataset~\citep{hani2020minneapple} as a well-curated testbed capturing dense layouts, heavy occlusion, and substantial appearance variation, we provide the first systematic evaluation of SAM3 against fine-tuned YOLO11 variants under controlled, reproducible conditions. Rather than reporting metrics alone, we analyze why these models diverge in behavior and how evaluation settings, especially IoU matching criteria, shape the perceived performance gap. A key methodological finding is that standard IoU thresholds (0.3–0.5) can distort results in dense scenes by penalizing masks with minor boundary deviations. Through systematic IoU sweeps, we identify 0.15 as a more appropriate threshold, reducing evaluation bias and enabling fair comparison. Our IoU sensitivity analysis reveals that SAM3 exhibits remarkable boundary quality stability (only 4-point F1 drop from IoU=0.10 to 0.50), whereas YOLO11 models show steep degradation (48-50 point drops), suggesting distinct trade-offs between detection completeness and mask precision.

\subsection{Contributions}
This work makes several key contributions to agricultural computer vision and instance segmentation research. First, we provide the first systematic comparison of zero-shot SAM3 against fine-tuned YOLO11 models specifically for agricultural instance segmentation, filling a critical gap in understanding how foundation models perform relative to specialized detectors in dense, occluded environments. Our comprehensive evaluation across three YOLO11 variants (2.5M to 27.6M parameters) reveals scaling laws and capacity trade-offs in small-data regimes typical of agricultural applications.

Second, we identify and resolve a critical evaluation issue where standard IoU thresholds mischaracterize model performance in dense scenes. Our IoU sensitivity analysis and threshold selection procedure provide a replicable methodology for researchers working with dense object detection tasks. We demonstrate the importance of aligning training and evaluation metrics, documenting how validation mAP$_{50}$ can appear inconsistent with test F1 scores when evaluation methodologies differ.

Third, we offer concrete deployment recommendations based on empirical trade-offs. For production systems requiring maximum accuracy, YOLO11 medium provides optimal performance at 72.2\% F1 with reasonable computational cost. For resource-constrained edge deployment, YOLO11 nano achieves 68.9\% F1 with 55$\times$ faster inference than SAM3. For zero-shot scenarios without labeled data, SAM3 provides immediate 59.8\% F1 functionality. These recommendations enable practitioners to make informed decisions based on their specific constraints and requirements.

\subsection{Objectives of this Study}
To make this contribution precise, we formulate the following objectives:

\begin{itemize}
    \item \textbf{Objective 1:} Quantitatively compare zero-shot SAM3 and fine-tuned YOLO11 variants on a dense, publicly available dataset (MinneApple~\citep{hani2020minneapple}), using consistent training, prompting, and evaluation protocols, in order to characterize the specialization versus generalization trade-off in instance segmentation.
    \item \textbf{Objective 2:} Systematically analyze the impact of IoU threshold selection and related evaluation choices on reported performance, and identify task-appropriate operating points for dense, overlapping instance segmentation that can generalize to similar real-world datasets.
    \item \textbf{Objective 3:} Study scaling behavior within the YOLO11 family by comparing nano, medium, and large variants, thereby elucidating how model capacity, accuracy, recall, precision, and computational cost interact in limited-data regimes.
    \item \textbf{Objective 4:} Provide empirically grounded guidance for practitioners on when to favor specialized fine-tuned models versus generalist foundation models, including concrete recommendations on model selection, configuration, and evaluation for MinneApple-like environments.
\end{itemize}

In the remainder of the paper, we build on these objectives to systematically explore, analyze, and report the capabilities and limitations of specialized versus generalist segmentation models in dense natural environments, using agricultural imagery as a reproducible and scientifically meaningful case study.

\begin{figure*}[h!]
\centering
\scriptsize

\newcommand{\iconRoot}{%
  \tikz[baseline=-0.5ex]\draw[fill=gray!60,draw=gray!80!black] (0,0) circle (0.6ex);}

\newcommand{\iconData}{%
  \tikz[baseline=-0.5ex]\draw[fill=purple!60,draw=purple!80!black]
    (-0.6ex,-0.35ex) -- (-0.6ex,0.35ex)
    arc (180:360:0.6ex and 0.2ex) -- (0.6ex,-0.35ex)
    arc (0:180:0.6ex and 0.2ex);}

\newcommand{\iconYolo}{%
  \tikz[baseline=-0.5ex]\draw[fill=blue!60,draw=blue!80!black,rounded corners=0.25ex]
    (-0.6ex,-0.4ex) rectangle (0.6ex,0.4ex);}

\newcommand{\iconSam}{%
  \tikz[baseline=-0.5ex]\draw[fill=orange!60,draw=orange!80!black]
    (0,0.6ex) -- (-0.6ex,-0.6ex) -- (0.6ex,-0.6ex) -- cycle;}

\newcommand{\iconEval}{%
  \tikz[baseline=-0.5ex]\draw[fill=green!60,draw=green!80!black] (0,0) circle (0.6ex);}

\newcommand{\iconSubData}{%
  \tikz[baseline=-0.5ex]\draw[fill=purple!50,draw=purple!70!black]
    (-0.5ex,-0.3ex) -- (-0.5ex,0.3ex)
    arc (180:360:0.5ex and 0.18ex) -- (0.5ex,-0.3ex)
    arc (0:180:0.5ex and 0.18ex);}

\newcommand{\iconSubYolo}{%
  \tikz[baseline=-0.5ex]\draw[fill=blue!50,draw=blue!70!black,rounded corners=0.2ex]
    (-0.5ex,-0.35ex) rectangle (0.5ex,0.35ex);}

\newcommand{\iconSubSam}{%
  \tikz[baseline=-0.5ex]\draw[fill=orange!50,draw=orange!70!black]
    (0,0.5ex) -- (-0.5ex,-0.5ex) -- (0.5ex,-0.5ex) -- cycle;}

\newcommand{\iconSubEval}{%
  \tikz[baseline=-0.5ex]\draw[fill=green!50,draw=green!70!black] (0,0) circle (0.5ex);}

\begin{forest}
for tree={
    grow=east,
    draw,
    rounded corners,
    line width=0.5pt,
    anchor=west,
    parent anchor=east,
    child anchor=west,
    edge={->, line width=0.6pt},
    font=\sffamily\tiny,
    minimum height=0.7em,
    s sep=3.0mm,
    l sep=4.0mm,
    text width=4.4cm,
    align=left,
    inner sep=2pt
}
[{\iconRoot\ \shortstack{Experimental methodology:\\SAM3 vs.\ YOLO11}},
  fill=gray!20,
  text width=3.0cm,
  align=center
  [{{\iconData\ \shortstack{Dataset Preparation\\(Sec.~\ref{sec:Dataset_Selection_and_Preparation})}}},
    fill=purple!15
    [{{\iconSubData\ MinneApple: 670 images, 28K instances}}]
    [{{\iconSubData\ PNG $\rightarrow$ YOLO polygon labels}}]
    [{{\iconSubData\ Split: Train 468, Val 100, Test 102}}]
  ]
  [{{\iconYolo\ \shortstack{YOLO11 Fine-Tuning\\(Sec.~\ref{sec:yolo11_finetuning})}}},
    fill=blue!15
    [{{\iconSubYolo\ 3 variants: n (2.5M), m (10.8M), l (27.6M)}}]
    [{{\iconSubYolo\ 100 epochs, AdamW, cosine LR}}]
    [{{\iconSubYolo\ Augmentations: mosaic=0.5, copy-paste=0.5, mix=0.3}}]
    [{{\iconSubYolo\ Monitored metrics: mAP$_{50}$, losses, precision/recall}}]
  ]
  [{{\iconSam\ \shortstack{SAM3 Zero-Shot Segmentation\\(Sec.~\ref{sec:sam3_zero_shot})}}},
    fill=orange!15
    [{{\iconSubSam\ $\sim$310M parameters, ViT-based encoder}}]
    [{{\iconSubSam\ Text prompts (e.g., ``apple'') via CLIP-style embedding}}]
    [{{\iconSubSam\ Inference at 1280$\times$960, confidence = 0.50}}]
  ]
  [{{\iconEval\ \shortstack{Evaluation Protocol\\(Sec.~\ref{sec:evaluation_metrics})}}},
    fill=green!15
    [{{\iconSubEval\ IoU-based matching (Eq.~\ref{eq:iou})}}]
    [{{\iconSubEval\ Precision, recall, F1 (Eqs.~\ref{eq:precision}--\ref{eq:f1})}}]
    [{{\iconSubEval\ IoU threshold $\tau = 0.15$ for dense fruit scenes}}]
    [{{\iconSubEval\ mAP$_{50}$ for training diagnostics}}]
    [{{\iconSubEval\ NMS and confidence threshold tuning}}]
  ]
  [{{\iconEval\ \shortstack{Ablation and Sensitivity Studies\\(Sec.~\ref{sec:ablation})}}},
    fill=red!15
    [{{\iconSubEval\ Confidence sweeps: $\theta \in [0.15, 0.40]$}}]
    [{{\iconSubEval\ Augmentation study: weak vs.\ strong pipelines}}]
    [{{\iconSubEval\ Dataset split balance and distribution checks}}]
    [{{\iconSubEval\ IoU sensitivity: $\tau \in [0.05, 0.50]$}}]
  ]
]
\end{forest}

\caption{Tree-style overview of the experimental methodology for comparing YOLO11 and SAM3. The diagram summarizes dataset preparation, YOLO11 fine-tuning, SAM3 zero-shot segmentation, evaluation protocol, and ablation studies, with icons highlighting the major components and section references guiding readers to detailed descriptions.}
\label{fig:method_overview}
\end{figure*}
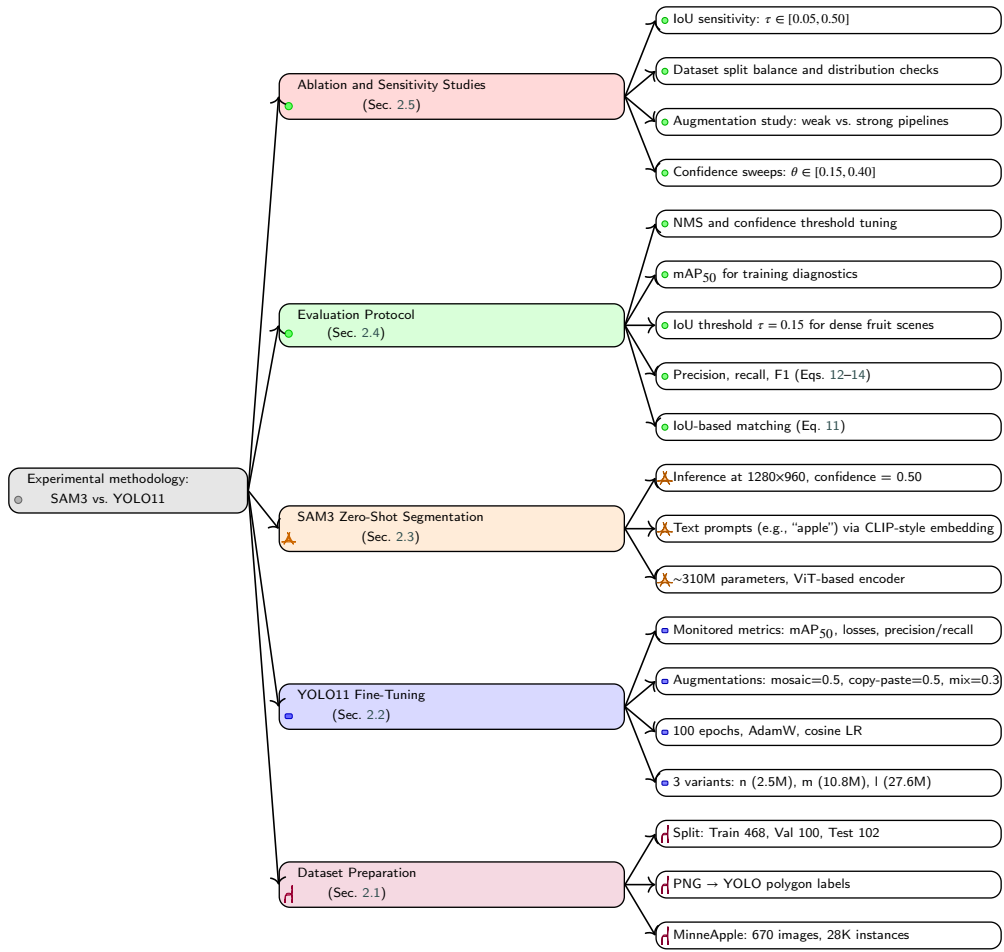

\section{Methodology}
\label{sec:Methodology}

Our experimental methodology consists of five main components as depicted in Figure \ref{fig:method_overview}: (i) dataset selection, preparation, and analysis; (ii) YOLO11 fine-tuning procedures; (iii) SAM3 zero-shot inference configuration; (iv) evaluation metrics, threshold selection, and (v) targeted ablation and sensitivity studies. This section details each component to ensure reproducibility and to make explicit all computations reported in the Results (Section~\ref{sec:results}) and Discussion (Section~\ref{sec:discussion}).
\subsection{Dataset Selection and Preparation}
\label{sec:Dataset_Selection_and_Preparation}

We adopt the MinneApple dataset~\citep{hani2020minneapple} as the benchmark for this comparative study. MinneApple was collected in commercial apple orchards at the University of Minnesota's Horticultural Research Center and provides dense instance-level annotations suitable for rigorous segmentation evaluation. The dataset contains 670 high-resolution RGB images with spatial resolution $1280 \times 960$ pixels, partitioned into train (468 images, 69.9\%), validation (100 images, 14.9\%), and test (102 images, 15.2\%) splits. In total, there are 28{,}179 individually annotated apple instances, with an average of 42.1 apples per training image, 45.3 per validation image, and 38.9 per test image.

The test split exhibits slightly lower instance density than the training and validation splits, making it marginally less challenging and suitable for assessing generalization. The dataset has several properties that are representative of dense, real-world environments: (i) high object density with substantial occlusion and overlapping fruits; (ii) variability in object size, color, and maturity stage; (iii) diverse environmental conditions (illumination, shadows, background clutter); and (iv) realistic image imperfections such as motion blur, variable focus, and partial occlusions.

\subsubsection{Data Format Conversion Pipeline}
\label{sec:data_conversion}

MinneApple provides instance masks as PNG images where each non-zero pixel value encodes a unique apple instance. This format is directly compatible with pixel-wise evaluation and foundation models such as SAM3, but YOLO11 requires polygon annotations in its native text format. We therefore implemented a conversion pipeline (\texttt{convert\_masks\_to\_yolo.py}) to transform instance masks into YOLO-style polygon labels.

For each image, the conversion proceeds as follows:
\begin{enumerate}
    \item Load the PNG instance mask, where each non-zero integer value $k$ corresponds to a distinct apple instance $I_k$.
    \item For each instance ID $k$, construct a binary mask $B_k(x,y)$ where
    \begin{equation}
        B_k(x,y) =
        \begin{cases}
            1, & \text{if pixel } (x,y) \text{ belongs to instance } I_k, \\
            0, & \text{otherwise}.
        \end{cases}
    \end{equation}
    \item Apply OpenCV's \texttt{findContours} with \texttt{RETR\_EXTERNAL} and \texttt{CHAIN\_APPROX\_SIMPLE} to extract the external contour of $B_k$.
    \item Simplify the contour using the Douglas–Peucker algorithm via \texttt{approxPolyDP} with tolerance
    \begin{equation}
        \epsilon_k = \alpha \, P_k,
    \end{equation}
    where $P_k$ is the perimeter (in pixels) of the original contour for instance $I_k$ and $\alpha = 0.001$ is a dimensionless tolerance parameter. The Douglas–Peucker approximation reduces the number of polygon vertices while preserving shape fidelity.
    \item Normalize polygon coordinates by the image width $W$ and height $H$:
    \begin{equation}
        \tilde{x}_i = \frac{x_i}{W}, \qquad
        \tilde{y}_i = \frac{y_i}{H},
    \end{equation}
    where $(x_i, y_i)$ are the original pixel coordinates of the $i$-th vertex and $(\tilde{x}_i, \tilde{y}_i)$ are the corresponding normalized coordinates.
    \item Write each instance as a single line in a YOLO label file:
    \begin{equation}
        \texttt{class\_id} \ \tilde{x}_1 \ \tilde{y}_1 \ \tilde{x}_2 \ \tilde{y}_2 \dots \tilde{x}_n \ \tilde{y}_n,
    \end{equation}
    where \texttt{class\_id} = 0 for apples (single class) and $(\tilde{x}_i, \tilde{y}_i)$ are the normalized polygon vertices.
\end{enumerate}

We manually verified the conversion quality by visual inspection of 50 randomly sampled images, overlaying the generated polygons on the original instance masks to confirm that the Douglas–Peucker simplification with $\alpha=0.001$ preserves boundary fidelity without introducing perceptible shape distortion.

\subsubsection{Dataset Statistics and Image Coverage}
\label{sec:dataset_stats}

We performed a split-wise statistical analysis to quantify dataset characteristics and verify the absence of harmful distribution shifts. Table~\ref{tab:dataset_stats} summarizes the key statistics.

\begin{table}[ht]
\centering
\caption{MinneApple dataset statistics across train, validation, and test splits.}
\label{tab:dataset_stats}
\begin{tabular}{lccc}
\toprule
\textbf{Metric} & \textbf{Train} & \textbf{Validation} & \textbf{Test} \\
\midrule
Number of Images & 468 & 100 & 102 \\
Total Apple Instances & 19{,}682 & 4{,}528 & 3{,}969 \\
Mean Instances/Image & 42.1 & 45.3 & 38.9 \\
Median Instances/Image & 39.0 & 42.5 & 37.0 \\
Min Instances/Image & 2 & 1 & 1 \\
Max Instances/Image & 123 & 119 & 117 \\
Image Coverage (\%) & 4.5 & 4.8 & 5.3 \\
\bottomrule
\end{tabular}
\end{table}

Image coverage measures the proportion of image pixels occupied by apple instances. For a split $S$ with $N_S$ images, width $W$, height $H$, and per-image apple masks $M_{i}(x,y)$ (union of all instance masks in image $i$), we define
\begin{equation}
    \text{Coverage}_S = 
    \frac{
        \sum_{i=1}^{N_S} \sum_{x=1}^{W} \sum_{y=1}^{H} \mathbb{1}\big(M_{i}(x,y) = 1\big)
    }{
        N_S \cdot W \cdot H
    } \times 100\%,
\end{equation}
where $\mathbb{1}(\cdot)$ is the indicator function. In the context of MinneApple, a pixel with $M_{i}(x,y)=1$ corresponds to any pixel belonging to at least one apple instance in image $i$. The similar coverage and instance distributions across splits confirm that the training, validation, and test sets are well-balanced; thus, performance differences across models primarily reflect model behavior rather than dataset artifacts.

\subsection{YOLO11 Fine-Tuning}
\label{sec:yolo11_finetuning}

We fine-tuned three YOLO11 segmentation variants, nano (YOLO11n-seg), medium (YOLO11m-seg), and large (YOLO11l-seg), to study how model capacity affects performance in dense scenes (Table~\ref{tab:yolo_variants}). All models were initialized from COCO-pretrained weights provided by Ultralytics~\citep{yolo11_2024} and trained on the MinneApple training split.

\subsubsection{Model Architectures and Capacity}
\label{sec:yolo_arch}

The three YOLO11 variants share the same architectural design (backbone, neck, and segmentation head) but differ in depth and width, resulting in different parameter counts and computational costs. Table~\ref{tab:yolo_variants} summarizes their characteristics.

\begin{table*}[ht]
\centering
\caption{YOLO11 segmentation variants: architecture specifications and computational requirements (640$\times$640 input).}
\label{tab:yolo_variants}
\begin{tabular}{lcccc}
\toprule
\textbf{Variant} & \textbf{Parameters} & \textbf{GFLOPs} & \textbf{Model Size} & \textbf{Speed (ms)} \\
\midrule
YOLO11n-seg & 2.5M & 10.4 & 5.7 MB & 45 \\
YOLO11m-seg & 10.8M & 65.3 & 21 MB & 128 \\
YOLO11l-seg & 27.6M & 132.6 & 54 MB & 217 \\
\bottomrule
\end{tabular}
\end{table*}

We use GFLOPs (billions of floating-point operations per forward pass) as a standard measure of computational cost. Let $C$ denote the computational cost in GFLOPs and let $F_1$ denote the dataset-level F1-score (Section~\ref{sec:prf_definition}). For later analysis, we define an efficiency metric
\begin{equation}
    E_{\text{F1}} = \frac{F_1}{C},
\end{equation}
which quantifies the segmentation performance obtained per unit of computation.

\subsubsection{Training Configuration and Hyperparameters}
\label{sec:yolo_training}

All YOLO11 variants were fine-tuned under a consistent training regime to enable fair comparison:

\begin{itemize}
    \item \textbf{Epochs:} 100 epochs with early stopping patience of 50 epochs, monitoring validation mask mAP at IoU=0.5 (mAP$_{50}$).
    \item \textbf{Batch size:} 16 images per batch (distributed across available GPUs).
    \item \textbf{Input resolution:} 640$\times$640 pixels, preserving aspect ratio with letterboxing.
    \item \textbf{Optimizer:} AdamW with initial learning rate $\eta_0 = 0.002$ and weight decay configured as in Ultralytics defaults.
    \item \textbf{Learning rate schedule:} Cosine annealing from $\eta_0$ to a minimum learning rate $\eta_{\min}$ over the course of training.
\end{itemize}

An extensive data augmentation pipeline is used to mitigate overfitting and enhance generalization in this limited-data regime. The final configuration includes:

\begin{itemize}
    \item \textbf{Mosaic:} with probability $p_{\text{mosaic}} = 0.5$, four images are combined into a single composite image.
    \item \textbf{Copy–paste:} with probability $p_{\text{cp}} = 0.5$, apple instances are pasted from other images onto the current image.
    \item \textbf{Mixup:} with probability $p_{\text{mixup}} = 0.3$, two images and their labels are blended.
    \item \textbf{Rotation:} random in-plane rotation in the range $\pm 15^\circ$.
    \item \textbf{Color jitter in HSV:} hue shift in $\pm 0.015$, saturation scaling up to $\pm 0.7$, and value scaling up to $\pm 0.4$.
\end{itemize}

This augmentation configuration evolved from an initial, weaker setup (mosaic=0.2, copy–paste=0.3) after we observed a validation–test performance gap. Strengthening augmentation improved validation mask mAP$_{50}$ and led to more consistent behavior across splits (see Section~\ref{sec:results_protocol}).

\subsubsection{Training Infrastructure and Monitoring}
\label{sec:yolo_infra}

Training was performed on ThunderCompute~\citep{thundercompute2025} cloud instances equipped with dual NVIDIA A100 GPUs (80\,GB VRAM each), 18 vCPUs, 144\,GB system RAM, and 1\,TB SSD storage, running Ubuntu. The Ultralytics Python package (version 8.0+) with PyTorch 2.0+ served as the training framework. Approximate wall-clock times for 100 epochs were 8 minutes for YOLO11n, 11 minutes for YOLO11m, and 15 minutes for YOLO11l, reflecting the increasing computational cost with model capacity.

\begin{figure*}[!htbp]
\centering
\includegraphics[width=\textwidth]{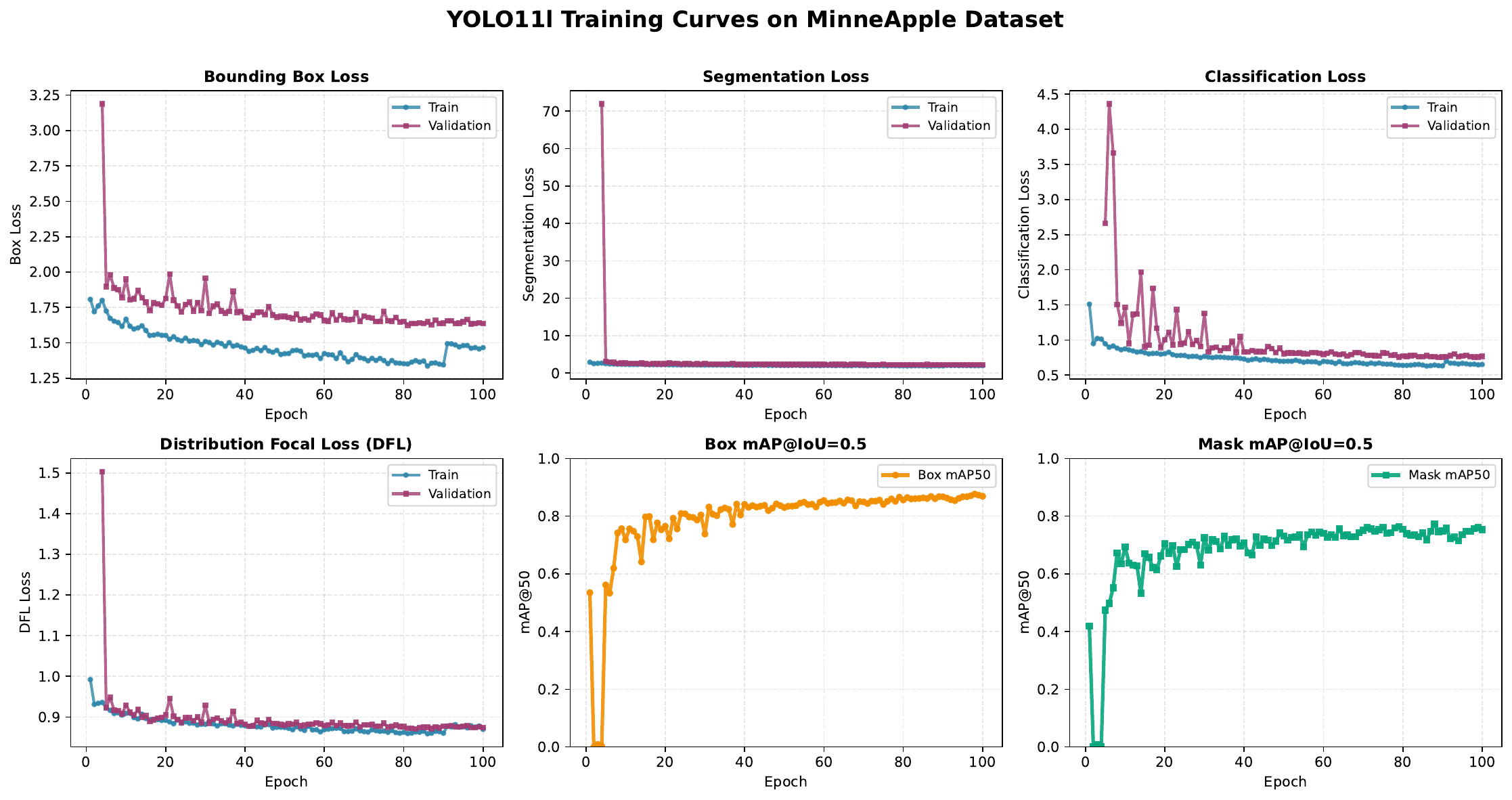}
\caption{Training curves for YOLO11l on the MinneApple dataset showing convergence of training and validation losses (box, segmentation, classification, and DFL) alongside mask and box mAP@IoU=0.5 performance metrics over 100 epochs. The model achieves 87.2\% validation mask mAP50, demonstrating effective learning under the enhanced augmentation regime (mosaic=0.5, copy-paste=0.5, mixup=0.3, rotation=$\pm$15°).}
\label{fig:training_curves}
\end{figure*}

During training, we monitored:
\begin{itemize}
    \item mask and box mAP$_{50}$ (mean Average Precision at IoU=0.5);
    \item precision and recall for boxes and masks;
    \item classification, bounding box regression, and mask losses.
\end{itemize}
Validation was performed after each epoch, and the checkpoint with highest validation mask mAP$_{50}$ was saved as \texttt{best.pt}, while the final epoch model was saved as \texttt{last.pt}. Figure~\ref{fig:training_curves} illustrates the training progression for YOLO11l.

\subsection{SAM3 Zero-Shot Segmentation}
\label{sec:sam3_zero_shot}

In contrast to YOLO11, which is fine-tuned on MinneApple, SAM3 is evaluated in its intended zero-shot configuration. We treat SAM3 as a generalist foundation model~\citep{kirillov2023segment,sam3_2024} and assess how far its pre-trained capabilities can go on MinneApple-like data without any adaptation.

\subsubsection{Model Architecture and Prompting}
\label{sec:sam3_arch_prompt}

SAM3 consists of three main components:
\begin{itemize}
    \item an image encoder (vision transformer) that maps an input image $I$ to a dense feature representation;
    \item a prompt encoder that maps prompts (points, boxes, text) into an embedding space;
    \item a mask decoder that fuses image and prompt embeddings to predict instance masks.
\end{itemize}

We use a purely textual prompt, the single-word phrase ``apple'', which is encoded using a CLIP-aligned text encoder. During inference, SAM3 produces a set of predicted masks
\begin{equation}
    \hat{\mathcal{M}} = \{ \hat{M}_j \}_{j=1}^{J},
\end{equation}
each associated with a confidence score $s_j \in [0,1]$. We apply a confidence threshold $\theta_{\text{SAM}}$ and retain only predictions with $s_j \geq \theta_{\text{SAM}}$:
\begin{equation}
    \hat{\mathcal{M}}_{\theta_{\text{SAM}}} = \{ \hat{M}_j \in \hat{\mathcal{M}} \mid s_j \geq \theta_{\text{SAM}} \}.
\end{equation}
Based on validation-set tuning (Section~\ref{sec:evaluation_metrics}), we set $\theta_{\text{SAM}} = 0.50$.

\subsubsection{Inference Configuration and Compute Profile}
\label{sec:sam3_inference}

SAM3 inference is conducted on full-resolution images ($1280 \times 960$) to preserve boundary detail. The model is loaded from the Hugging Face Model Hub (\texttt{facebook/sam3}) using the \texttt{transformers} library, and executed with automatic mixed precision (AMP) on NVIDIA A100 GPUs (Section~\ref{sec:yolo_infra}).

Let $T_i$ denote the inference time for image $i$ and $N$ the number of images in the test set. The total inference time is
\begin{equation}
    T_{\text{total}} = \sum_{i=1}^{N} T_i,
\end{equation}
and the average per-image runtime is
\begin{equation}
    \bar{T} = \frac{1}{N} \sum_{i=1}^{N} T_i.
\end{equation}
These definitions are applied consistently to both SAM3 and YOLO11 variants, enabling fair comparison of computational efficiency (Section~\ref{sec:results_efficiency}). For SAM3, the average per-image runtime is on the order of seconds, with GPU memory usage of approximately 8\,GB.

\subsection{Evaluation Metrics and Threshold Selection}
\label{sec:evaluation_metrics}

Accurate and fair evaluation is central to our study. We base our evaluation on instance-level matching using Intersection over Union (IoU), followed by computation of precision, recall, and F1-score. We also clarify the meaning of true positives, false positives, and false negatives in the specific context of MinneApple apples, and define additional derived quantities reported in the Results and Discussion.

\subsubsection{Intersection over Union and Instance Matching}
\label{sec:iou_definition}

For a predicted mask $M_p$ and a ground-truth mask $M_g$ (both subsets of the image lattice), IoU is defined as
\begin{equation}
    \text{IoU}(M_p, M_g) = 
    \frac{\left| M_p \cap M_g \right|}{\left| M_p \cup M_g \right|},
    \label{eq:iou}
\end{equation}
where $|\cdot|$ denotes the number of pixels in the set. Given an IoU threshold $\tau \in (0,1]$, a predicted apple mask $M_p$ is considered a \emph{true positive} (TP) if:
\begin{enumerate}
    \item there exists a ground-truth apple instance mask $M_g$ such that $\text{IoU}(M_p, M_g) \geq \tau$, and
    \item $M_g$ has not already been matched by another prediction with higher confidence.
\end{enumerate}
In MinneApple, a TP therefore corresponds to a predicted apple instance that sufficiently overlaps a specific annotated apple instance.

A predicted mask that does not satisfy these conditions for any ground-truth apple is counted as a \emph{false positive} (FP); in MinneApple, an FP is either a spurious apple prediction (background or foliage) or a duplicate prediction for an apple that is already matched. Conversely, any ground-truth apple mask that is not matched by any prediction is a \emph{false negative} (FN), representing a missed apple instance.

\subsubsection{Precision, Recall, and F1-Score (Dataset-Level)}
\label{sec:prf_definition}

Let $\text{TP}$, $\text{FP}$, and $\text{FN}$ denote the total counts across the entire test set (aggregated over all images). Precision, recall, and F1-score are defined as
\begin{equation}
    \text{Precision}(\tau) = \frac{\text{TP}(\tau)}{\text{TP}(\tau) + \text{FP}(\tau)},
    \label{eq:precision}
\end{equation}
\begin{equation}
    \text{Recall}(\tau) = \frac{\text{TP}(\tau)}{\text{TP}(\tau) + \text{FN}(\tau)},
    \label{eq:recall}
\end{equation}
\begin{equation}
    F_1(\tau) = 2 \cdot \frac{\text{Precision}(\tau) \cdot \text{Recall}(\tau)}{\text{Precision}(\tau) + \text{Recall}(\tau)}.
    \label{eq:f1}
\end{equation}
Here, $\text{TP}(\tau)$, $\text{FP}(\tau)$, and $\text{FN}(\tau)$ emphasize that these counts depend on the IoU threshold $\tau$ used for matching. In the context of MinneApple, $\text{TP}(\tau)$ is the number of correctly detected apple instances across all test images at threshold $\tau$, $\text{FP}(\tau)$ is the number of predicted apple instances that do not correspond to any annotated apple, and $\text{FN}(\tau)$ is the number of annotated apples that are missed.

For image-level analysis, let $F_{1,i}(\tau)$ denote the F1-score computed for image $i$ alone. The mean image-level F1-score is then
\begin{equation}
    \overline{F_1}_{\text{image}}(\tau) = \frac{1}{N} \sum_{i=1}^{N} F_{1,i}(\tau),
    \label{eq:mean_image_f1}
\end{equation}
where $N$ is the number of images in the test set. This metric emphasizes per-image performance variability, which is particularly relevant for dense scenes with varying apple counts.

\subsubsection{Mean Average Precision at IoU=0.5}
\label{sec:map_definition}

During YOLO11 training, we monitor mask and box mean Average Precision at IoU=0.5 (mAP$_{50}$). For a binary class setting (apples vs.\ background) and fixed IoU threshold $\tau = 0.5$, the average precision (AP) is defined as the integral over the precision–recall curve
\begin{equation}
    \text{AP}_{50} = \int_{0}^{1} P(R) \, dR,
\end{equation}
where $P(R)$ is precision as a function of recall $R$, approximated numerically as the area under the discretized precision–recall curve. With a single class, mAP$_{50}$ reduces to AP$_{50}$. This training-time metric is used for model selection (Section~\ref{sec:yolo_training}), whereas $F_1(\tau)$ at $\tau = 0.15$ is the primary metric for test-set comparison.

\subsubsection{IoU Threshold Sensitivity and Choice of $\tau=0.15$}
\label{sec:iou_choice}

Standard segmentation benchmarks commonly adopt $\tau = 0.5$ or report mAP over a range of thresholds (e.g., 0.5:0.05:0.95). However, in dense scenes with significant occlusion and boundary ambiguity, such thresholds can be overly strict. In our experiments, initial evaluations at higher thresholds (e.g., $\tau=0.30$) revealed large discrepancies between training-time mAP$_{50}$ and test-set F1-scores.

To analyze this effect systematically, we consider a set of IoU thresholds
\begin{equation}
    \mathcal{T} = \{\tau_1, \tau_2, \dots, \tau_K\} \subset (0,1],
\end{equation}
and examine the function
\begin{equation}
    F_1 : \mathcal{T} \rightarrow [0,1], \quad \tau \mapsto F_1(\tau),
\end{equation}
where $F_1(\tau)$ is defined in \eqref{eq:f1}. For dense MinneApple scenes, we empirically observe that $F_1(\tau)$ is a monotonically decreasing function of $\tau$, reflecting the increasing strictness of the matching criterion. Visual inspection of apple clusters confirms that many predictions rejected at higher $\tau$ are semantically correct but exhibit small boundary deviations.

Given the high density and frequent occlusion in MinneApple, we select $\tau = 0.15$ as the primary evaluation threshold. This value requires substantial spatial overlap between prediction and ground truth while accommodating annotation uncertainty and boundary ambiguity. All headline metrics in Table~\ref{tab:main_results} are therefore computed at $\tau = 0.15$.

\subsubsection{Confidence Threshold Selection and NMS}
\label{sec:conf_nms}

In addition to IoU thresholding, we must select a confidence threshold $\theta$ to filter low-confidence predictions. Let $s(M_p)$ denote the confidence score associated with a predicted mask $M_p$. We define the set of retained predictions as
\begin{equation}
    \hat{\mathcal{M}}_{\theta} = \{ M_p \mid s(M_p) \geq \theta \}.
\end{equation}
For YOLO11 variants, we perform a grid search over a finite set $\Theta = \{\theta_1,\dots,\theta_L\}$ on the validation set and select
\begin{equation}
    \theta_{\text{YOLO}}^{\ast} = \arg\max_{\theta \in \Theta} F_{1,\text{val}}(\theta),
\end{equation}
where $F_{1,\text{val}}(\theta)$ is the validation-set F1-score obtained when using confidence threshold $\theta$. The selected $\theta_{\text{YOLO}}^{\ast}$ is then fixed for test-set evaluation. For SAM3, the same principle is applied (with its own candidate set $\Theta_{\text{SAM}}$), yielding $\theta_{\text{SAM}} = 0.50$ as described above.

To reduce duplicate predictions, we apply Non-Maximum Suppression (NMS) to YOLO11 outputs. Let $\tau_{\text{NMS}} \in (0,1]$ denote the NMS IoU threshold. NMS constructs a subset $\hat{\mathcal{M}}_{\theta_{\text{YOLO}}^{\ast},\text{NMS}} \subset \hat{\mathcal{M}}_{\theta_{\text{YOLO}}^{\ast}}$ such that no two masks in $\hat{\mathcal{M}}_{\theta_{\text{YOLO}}^{\ast},\text{NMS}}$ exceed IoU $\tau_{\text{NMS}}$, while preferring higher-confidence masks. SAM3’s instance generation mechanism is inherently non-redundant because its concept-conditioned decoder produces a single, semantically consolidated mask for each distinct concept rather than multiple overlapping proposals. By aligning vision and language features during decoding, SAM3 suppresses duplicate hypotheses at the representation level, enabling it to resolve instance boundaries through concept-level fusion rather than post-hoc filtering. As a result, the model minimizes repeated detections internally and eliminates the need for an explicit non-maximum suppression (NMS) step.

\subsection{Ablation and Sensitivity Studies}
\label{sec:ablation}

Several ablation and sensitivity analyses were performed to methodologically support the results and conclusions reported in Section~\ref{sec:results_protocol} and Section~\ref{sec:discussion_iou}:

\begin{itemize}
    \item \textbf{Confidence threshold sweep:} For each YOLO11 variant, we evaluated $F_1(\tau)$ at a fixed IoU threshold (e.g., $\tau = 0.30$) across multiple confidence thresholds $\theta \in \Theta$ and observed only modest variation, indicating that poor performance at high $\tau$ is driven primarily by IoU strictness rather than miscalibrated confidence thresholds.
    
    \item \textbf{Augmentation strength and overfitting check:} Initially observing a gap between validation mAP$_{50}$ (85.6\%) and test $F_1$ (53.5\% at $\tau=0.3$), we hypothesized model overfitting. To test this, we retrained YOLO11l with substantially stronger augmentation (increased mosaic from 0.2 to 0.5, copy-paste from 0.3 to 0.5, and added mixup at 0.3). The result: validation mAP$_{50}$ improved to 87.2\%, yet test $F_1$ remained unchanged at 53.5\%. This ruled out overfitting as the root cause, since a truly overfit model would show worse validation performance with stronger regularization, not better. The discrepancy persisted even on the validation set itself when evaluated with our custom IoU-based metrics, revealing that the gap was methodological rather than a generalization failure (Section~\ref{sec:Augmentation_and_Overfitting_Check}).
    
    \item \textbf{Dataset distribution analysis:} As detailed in Table~\ref{tab:dataset_stats}, the splits are well-balanced, and the test set is slightly less dense (38.9 vs. 45.3 apples/image in validation), indicating that dataset difficulty differences do not explain the observed performance gaps.
    
    \item \textbf{IoU sensitivity sweep and threshold discovery:} We systematically swept $\tau$ from 0.1 to 0.5 and observed that at $\tau=0.1$, YOLO11l achieved $F_1=75.9\%$, closely matching its training mAP$_{50}$ of 75.4\%. However, at $\tau=0.3$, performance collapsed to $F_1=44.4\%$, and at $\tau=0.5$, it further dropped to 19.4\%. This steep performance cliff revealed that strict IoU thresholds ($\tau \geq 0.3$) are unsuitable for dense orchard scenes where small boundary misalignments in tightly packed instances disproportionately reduce overlap scores. Based on this analysis, we adopted $\tau=0.15$ as a balanced threshold that reflects real-world detection quality while accounting for the inherent challenges of dense instance segmentation.
\end{itemize}

Together, these studies establish that evaluation methodology particularly IoU threshold selection plays a central role in how YOLO11 and SAM3 appear to perform on MinneApple-like datasets, and they justify the protocols adopted for all subsequent experiments.

\section{Results}
\label{sec:results}

This section reports the comparative performance of SAM3 in zero-shot mode and three fine-tuned YOLO11 variants on the MinneApple test set. We focus on instance-level and pixel-level segmentation accuracy, computational efficiency, error patterns, and the impact of IoU threshold selection in reporting the performance of these models. Interpretation of broader implications and deployment recommendations is deferred to the Discussion section.

\subsection{Overall Performance Comparison}
\label{sec:results_overall}

\begin{figure*}[!htbp]
\centering
\includegraphics[width=0.8\textwidth]{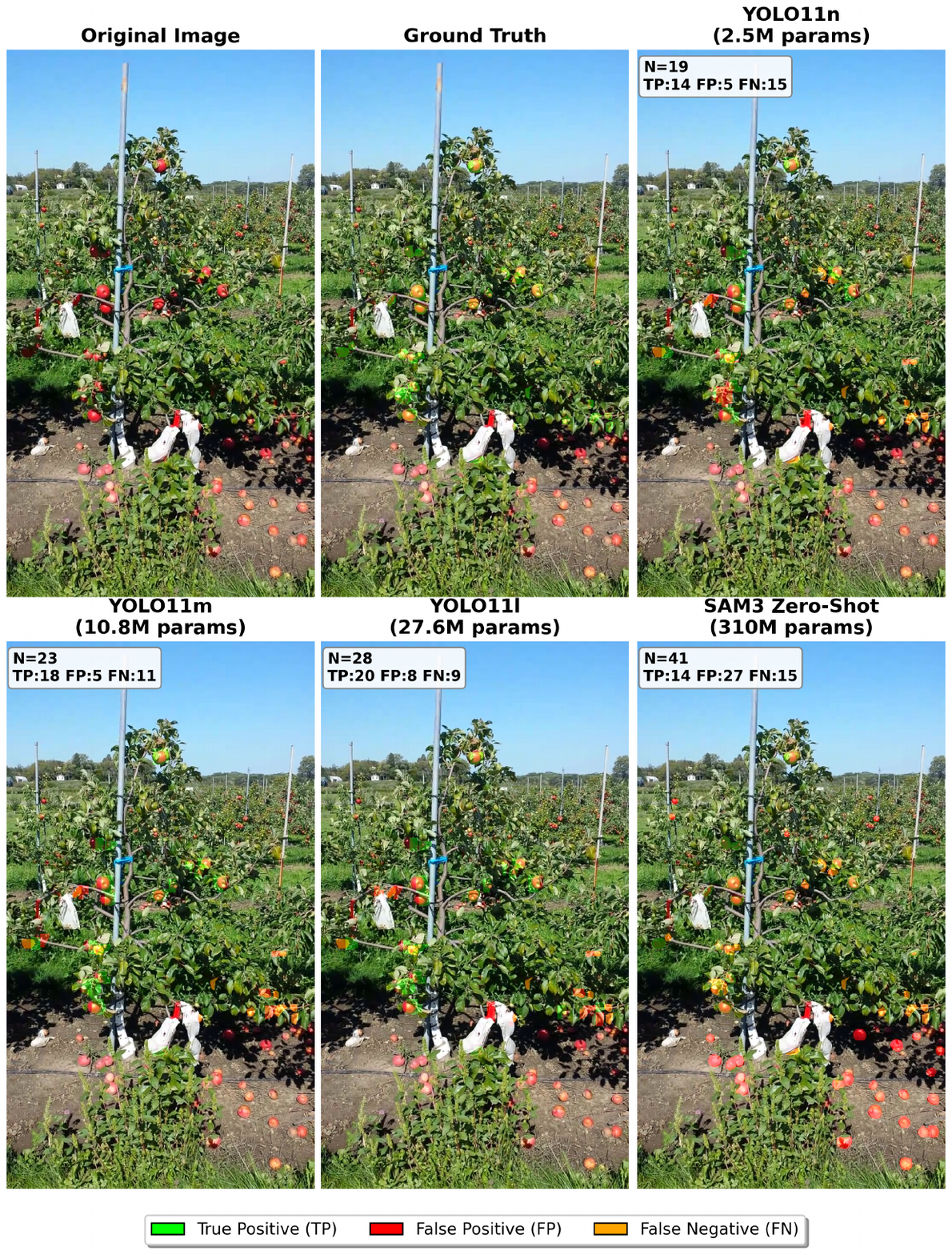}
\caption{Qualitative comparison on a representative test image. Top row: original, ground truth, YOLO11n. Bottom row: YOLO11m, YOLO11l, SAM3. True positives (green), false positives (red), false negatives (orange). YOLO11 models show higher precision with fewer false positives, while SAM3 exhibits more over-segmentation. Additional high-resolution examples available at~\href{https://github.com/Applied-AI-Research-Lab/Segment-Anything-Model-SAM3-Zero-Shot-Segmentation-Against-Fine-Tuned-YOLO-Detectors/blob/main/figures/fig_qualitative_grid.pdf}{\texttt{GitHub}}.}
\label{fig:qualitative_grid}
\end{figure*}

Table~\ref{tab:main_results} summarizes the primary instance segmentation results for all four models on the MinneApple test set. All metrics are computed at IoU\,=\,0.15, as motivated in Section~\ref{sec:evaluation_metrics}. Figure~\ref{fig:qualitative_grid} provides visual examples of these quantitative trends across different scene difficulties.

\begin{table*}[ht]
\centering
\caption{Instance segmentation performance on the MinneApple test set (IoU\,=\,0.15). TP, FP, and FN are reported as rates relative to the total 3,970 ground-truth instances.}
\label{tab:main_results}
\begin{tabular}{lccccccc}
\toprule
\textbf{Model} & \textbf{F1 (\%)} & \textbf{Precision (\%)} & \textbf{Recall (\%)} & \textbf{TP (\%)} & \textbf{FP (\%)} & \textbf{FN (\%)} & \textbf{Params} \\
\midrule
SAM3 Zero-Shot & 59.8 & 55.4 & 64.9 & 64.9 & 52.2 & 35.1 & 310M \\
YOLO11n (fine-tuned) & 68.9 & 78.6 & 61.3 & 61.3 & 16.7 & 38.7 & 2.5M \\
YOLO11m (fine-tuned) & \textbf{72.2} & 77.2 & 67.8 & 67.8 & 20.1 & 32.2 & 10.8M \\
YOLO11l (fine-tuned) & 71.9 & 75.2 & \textbf{68.8} & \textbf{68.8} & 22.7 & \textbf{31.2} & 27.6M \\
\bottomrule
\end{tabular}
\end{table*}

The fine-tuned YOLO11 models clearly outperform zero-shot SAM3 in instance-level F1-score. YOLO11m achieves the highest F1 at 72.2\%, a 12.4 percentage point improvement over SAM3 (59.8\%). YOLO11l attains the highest recall at 68.8\%, correctly detecting 2,733 of 3,970 ground-truth instances, i.e., 156 more instances than SAM3. Even the smallest YOLO11n model, with 124$\times$ fewer parameters than SAM3, exceeds the foundation model by 9.1 F1 points.

At the same time, SAM3’s 59.8\% F1 in pure zero-shot mode remains noteworthy: it is obtained without any task-specific training, fine-tuning, or adaptation to MinneApple, using only the single-word prompt ``apple''. The model still detects 2,577 apple instances in dense, occluded orchard scenes, indicating that generalist foundation models can reach performance levels within roughly 9--12 points of specialized detectors on this task.

The precision–recall profiles reveal distinct behaviors. SAM3 shows relatively balanced precision (55.4\%) and recall (64.9\%), but both are consistently lower than those of the fine-tuned models. All YOLO11 variants exhibit high precision (75 - 79\%) with moderate recall (61--69\%), acting as more conservative detectors that generate fewer false positives but may miss difficult or ambiguous instances. Among the YOLO11 variants, YOLO11m provides the best overall trade-off, slightly surpassing YOLO11l in F1 (72.2\% vs.\ 71.9\%) while using fewer parameters. YOLO11l’s higher recall but lower precision suggests that increasing model capacity helps recover more challenging instances at the cost of additional false positives.

\subsection{Comparison with prior Apple Detection Studies}
It is important to contextualize these results relative to the broader literature on YOLO-based fruit detection in orchards, where precision and recall values in the 90--95\% range are commonly reported~\citep{bargoti2017deep,tian2019apple,koirala2019deep}. Several methodological factors contribute to the apparent gap between our results and those studies. First, many prior works evaluate on officially curated test sets from benchmarks (such as the original MinneApple test set or similar datasets) that are specifically designed for leaderboard comparison and typically provide optimized class balance, lower scene complexity, or task-specific adjustments. In contrast, we chose \emph{not} to use the original MinneApple benchmark test set (which does not include public masks and is reserved for competition purposes). Instead, we extracted all labeled data from the training partition, then randomly split it into new training (69.9\%), validation (14.9\%), and test (15.2\%) subsets. This decision ensures reproducibility and full transparency of mask annotations for all splits, but it also means our test set may contain more challenging or ambiguous scenes than officially optimized benchmarks. Second, this study prioritizes \emph{fair comparison between specialized and generalist models} rather than maximizing absolute YOLO11 performance. We did not perform exhaustive hyperparameter tuning (e.g., searching over learning rates, batch sizes, advanced augmentation schedules, or architectural modifications) that would be standard practice when targeting state-of-the-art results. Our training configuration is representative but deliberately conservative to maintain consistency across model variants and to focus experimental effort on the comparative evaluation of YOLO11 versus SAM3. Third, our evaluation protocol uses IoU\,=\,0.15 for matching predictions to ground truth (Section~\ref{sec:evaluation_metrics}), which is more lenient than the standard mAP$_{50}$ (IoU\,=\,0.5) but still strict enough to penalize boundary imprecision. Studies reporting mAP$_{50}$ or mAP$_{50\text{--}95}$ often benefit from aggregation over multiple thresholds or optimized post-processing (e.g., test-time augmentation, ensemble models, threshold tuning on validation sets), which we do not apply here. Finally, the inherent density and occlusion in MinneApple (38.9--45.3 apples per image) present challenges that may not be equally pronounced in other datasets. In summary, the precision and recall values reported here (75--79\% precision, 61--69\% recall for YOLO11) reflect a deliberate trade-off: reproducibility, transparency, and methodological rigor for comparative analysis, rather than leaderboard-optimized performance. This approach ensures that the observed performance gaps between YOLO11 and SAM3 are attributable to intrinsic model characteristics and not artifacts of aggressive tuning or benchmark-specific engineering.

\subsection{Computational Efficiency Analysis}
\label{sec:results_efficiency}

Table~\ref{tab:efficiency} reports the computational characteristics of the four models, including parameter count, average inference time per image, GPU memory usage, and an F1-per-GFLOPs efficiency metric.

\begin{table*}[ht]
\centering
\caption{Computational efficiency comparison across models. Inference time is measured per image.}
\label{tab:efficiency}
\begin{tabular}{lcccc}
\toprule
\textbf{Model} & \textbf{Parameters} & \textbf{Inference Time (ms)} & \textbf{GPU Memory (GB)} & \textbf{F1 per GFLOPs} \\
\midrule
SAM3 Zero-Shot & 310M & 2500 & 8.0 & -- \\
YOLO11n & 2.5M & 45 & 2.0 & 6.62 \\
YOLO11m & 10.8M & 128 & 3.5 & 1.11 \\
YOLO11l & 27.6M & 217 & 4.0 & 0.54 \\
\bottomrule
\end{tabular}
\end{table*}

\begin{figure*}[!htbp]
\centering
\includegraphics[width=\textwidth]{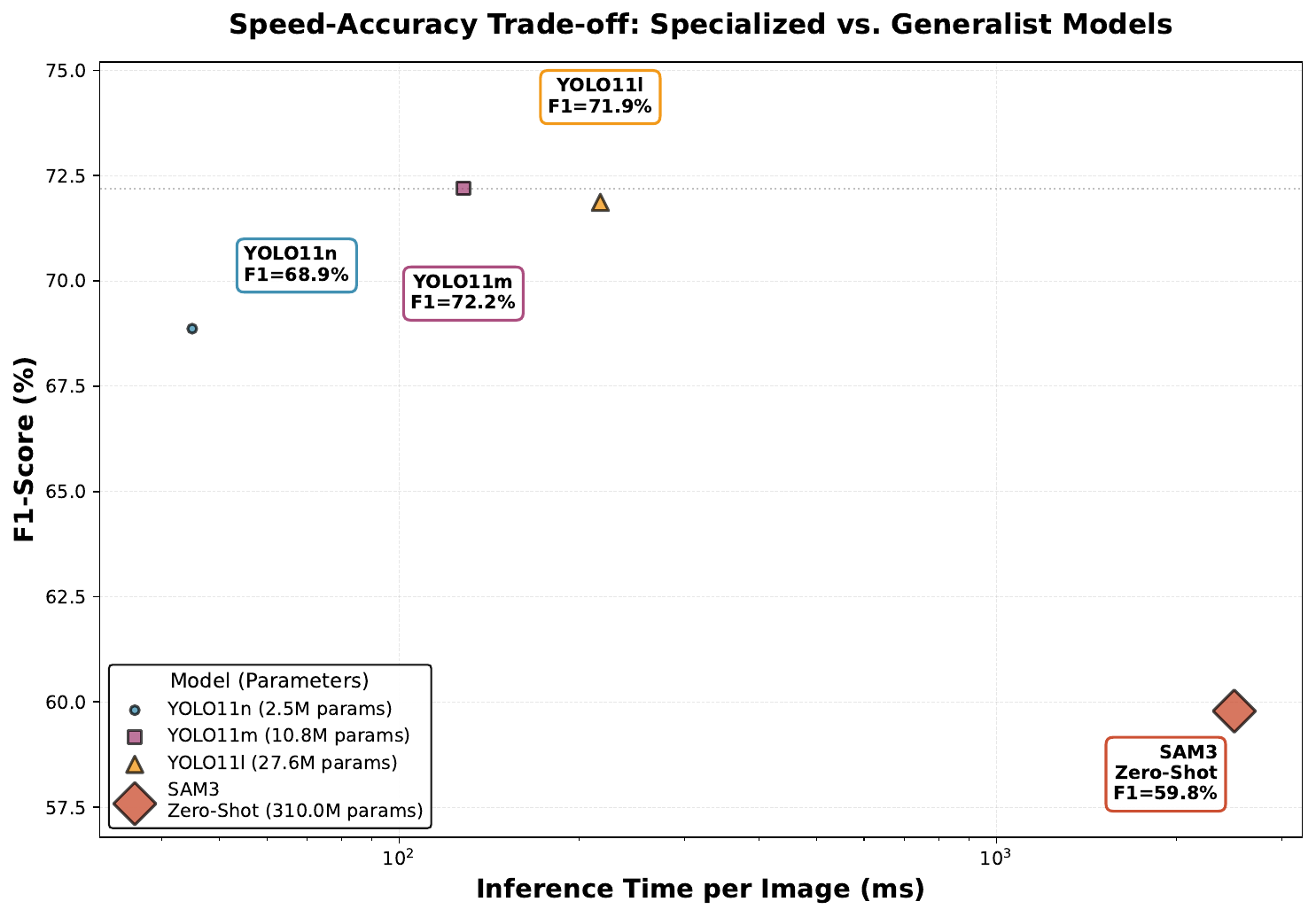}
\caption{Speed–accuracy trade-off comparing specialized fine-tuned YOLO11 models and zero-shot SAM3 on MinneApple test set. Each model is represented as a point in F1-score vs.\ inference-time space, with marker size proportional to parameter count (2.5M for YOLO11n to 310M for SAM3). The log-scale x-axis highlights the dramatic computational efficiency advantage of specialized models: YOLO11n achieves 68.9\% F1 in 45\,ms (55$\times$ faster than SAM3), while YOLO11m reaches the highest F1 of 72.2\% in 128\,ms (19.5$\times$ faster). SAM3's zero-shot performance (59.8\% F1, 2,500\,ms) demonstrates that despite using 124$\times$ more parameters than YOLO11n and requiring substantially longer inference time, the generalist foundation model underperforms all fine-tuned variants. The figure illustrates that task-specific fine-tuning yields simultaneous improvements in both accuracy and computational efficiency for dense instance segmentation.}
\label{fig:speed_accuracy}
\end{figure*}

The YOLO11 family exhibits a clear advantage in runtime and memory footprint compared to SAM3 (Figure~\ref{fig:speed_accuracy}). YOLO11n, the smallest variant, processes images approximately 55$\times$ faster than SAM3 (45\,ms vs.\ 2,500\,ms) while also achieving higher F1. Even YOLO11l, the largest YOLO11 model, remains 11.5$\times$ faster than SAM3, with lower memory consumption (4.0\,GB vs.\ 8.0\,GB). This result indicates that, under the evaluated conditions, specialization yields both better accuracy and substantially lower computational cost.

The F1-per-GFLOPs metric further quantifies efficiency. YOLO11n achieves the highest efficiency at 6.62 F1 points per GFLOP, making it attractive for resource-constrained or real-time deployments. As model capacity increases from YOLO11n to YOLO11l, absolute accuracy improves but efficiency diminishes, reflecting typical scaling trade-offs. SAM3, although not directly comparable via this metric in our setup, requires significantly more compute to reach a lower F1 score.

\subsection{Error Analysis and Failure Modes}
\label{sec:results_error}

The confusion statistics in Table~\ref{tab:main_results} reveal distinct error patterns across models. SAM3 produces 2,074 false positives, substantially more than any YOLO11 variant (663--903), indicating a tendency to over-segment scenes. Qualitative inspection suggests that these false positives often correspond to apple-like regions, partial views, or background structures (e.g., leaves, branches) that share visual characteristics with target instances. At the same time, SAM3 also incurs 1,393 false negatives, missing a non-trivial number of apples, especially when they are heavily occluded, partially visible, or poorly contrasted against the background.

\begin{figure*}[!htbp]
\centering
\includegraphics[width=\textwidth]{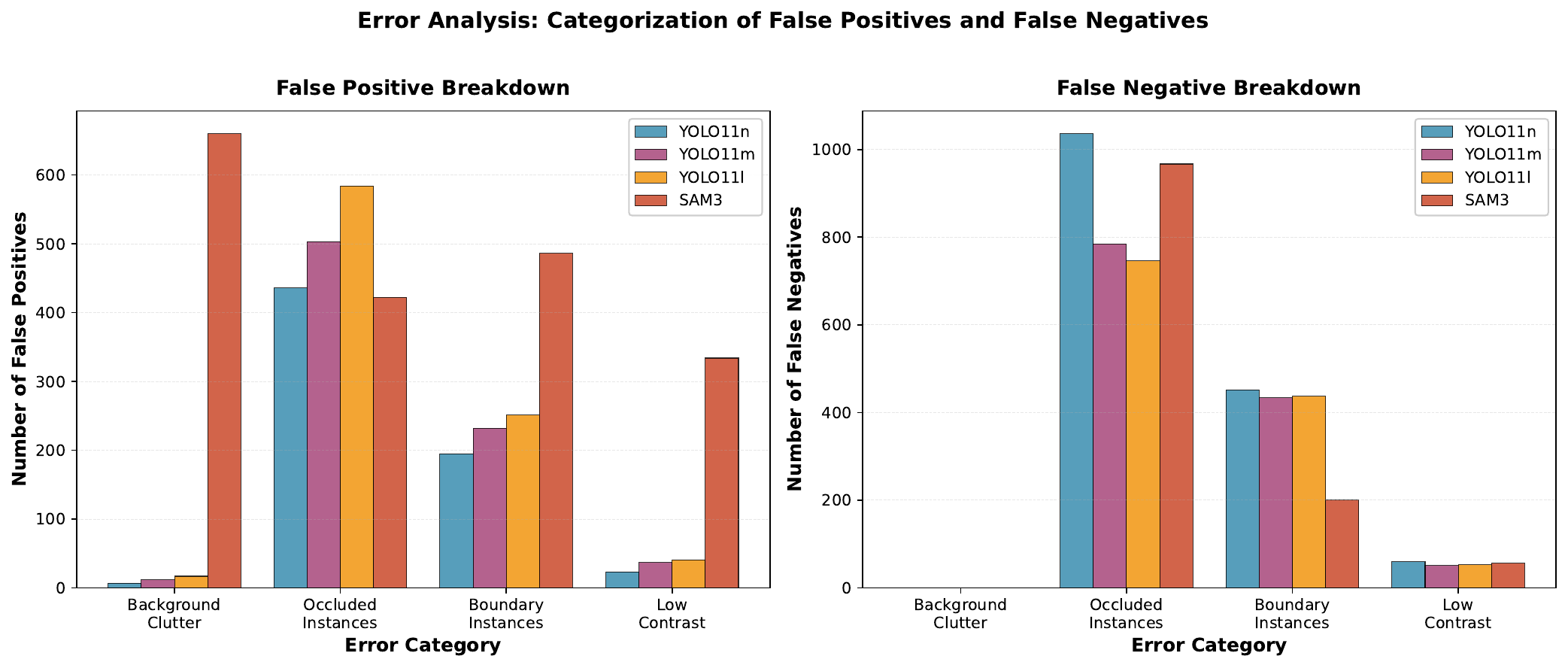}
\caption{Error breakdown analysis categorizing false positives (left) and false negatives (right) for each model into four interpretable groups. Categories are determined automatically using heuristic rules applied to spatial properties and image context, without manual annotation: \textit{boundary instances} are within 50 pixels of image edges; \textit{low-contrast apples} exhibit local standard deviation below 30 in their surrounding region; \textit{background clutter} (FP only) are isolated predictions with no ground-truth instances within a 100-pixel radius; and \textit{occluded instances} are in high-density regions with five or more ground-truth apples within a 200-pixel neighborhood. YOLO11 models (n/m/l) fail predominantly on occluded instances (60--67\% of all errors), reflecting the challenge of dense orchard scenes, followed by boundary instances (195--251 FPs, 434--452 FNs). In contrast, SAM3 exhibits a distinct error profile: 660 background clutter false positives (34.7\% of FPs) indicate poor specificity in distinguishing apples from non-apple regions, while 967 occluded false negatives (79.1\% of FNs) reveal difficulty with heavily overlapping instances. Low-contrast failures are relatively rare across all models (23--60 errors), suggesting that color and texture cues remain sufficiently discriminative even under challenging lighting. The automated categorization provides actionable insights into failure modes: YOLO11 models require improved handling of occlusion and boundary effects, whereas SAM3's zero-shot approach struggles fundamentally with scene specificity and dense instance disambiguation.}
\label{fig:error_breakdown}
\end{figure*}

In contrast, YOLO11 models consistently exhibit fewer false positives (663 for YOLO11n, 797 for YOLO11m, and 903 for YOLO11l), consistent with their higher precision. However, they produce more false negatives overall (1,537, 1,277, and 1,237, respectively), reflecting a more conservative detection behavior. These false negatives are often associated with challenging conditions such as strong occlusion, extreme lighting, or small object scale near image boundaries. As model capacity increases from YOLO11n to YOLO11l, false negatives decrease (1,537 $\rightarrow$ 1,237), indicating that larger models recover more difficult instances, albeit with a modest increase in false positives. Figure~\ref{fig:error_breakdown} categorizes these errors into interpretable failure modes.

\subsection{Performance Across IoU Thresholds}
\label{sec:results_iou}

\begin{figure*}[!htbp]
\centering
\includegraphics[width=\textwidth]{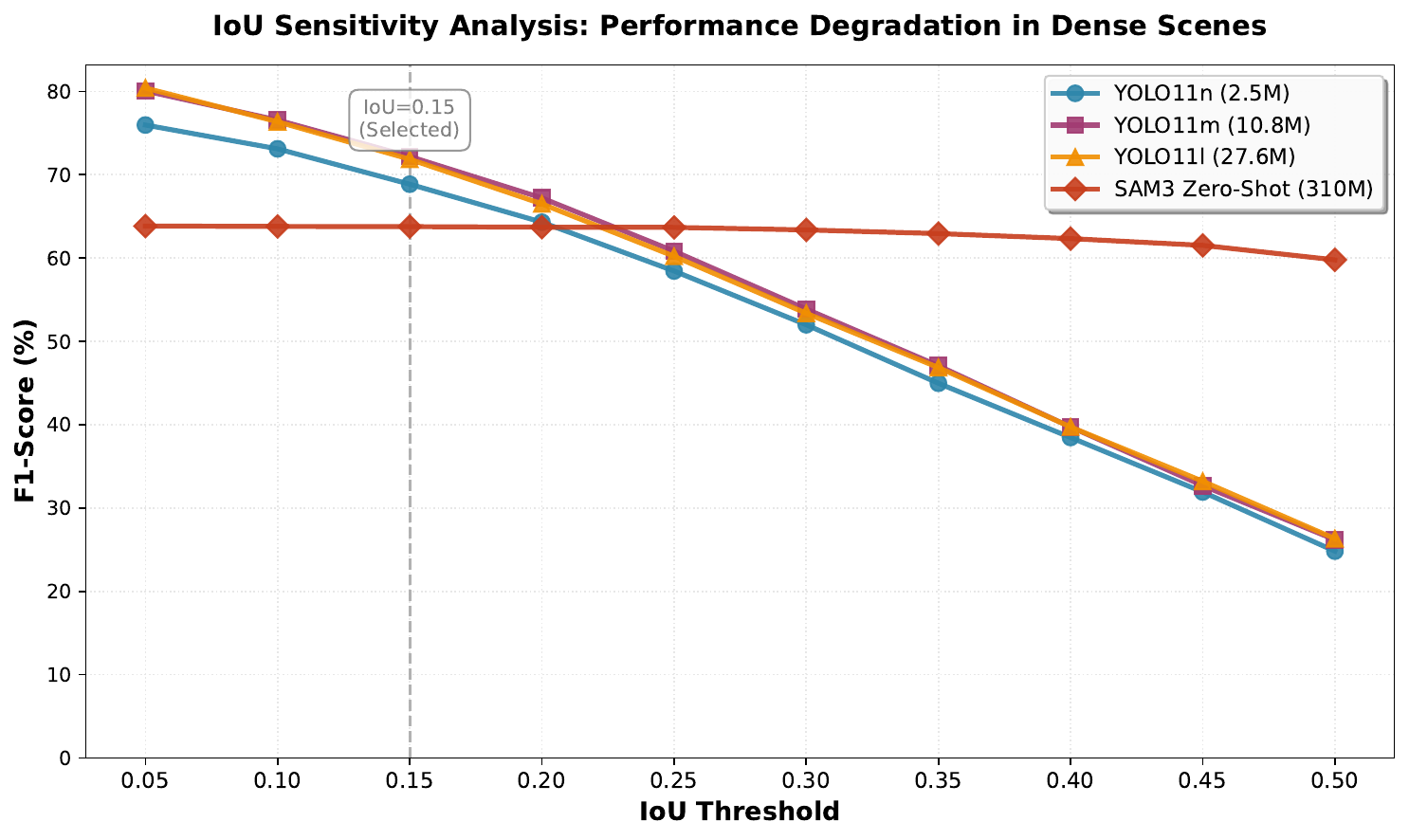}
\caption{IoU sensitivity analysis showing F1-score as a function of IoU matching threshold from 0.05 to 0.50 for all four models. YOLO11 variants exhibit steep performance degradation: YOLO11n drops 48.3 points (73.1\% to 24.8\%), YOLO11m drops 50.4 points (76.6\% to 26.2\%), and YOLO11l drops 50.1 points (76.4\% to 26.3\%) between IoU=0.10 and 0.50. In striking contrast, SAM3 shows remarkable stability with only a 4.0-point drop (63.8\% to 59.8\%) across the same range, demonstrating 12-fold lower sensitivity. The vertical dashed line marks IoU=0.15, our selected evaluation threshold for dense scenes. This stability suggests SAM3 produces higher-quality mask boundaries despite lower absolute F1 scores at lenient thresholds, revealing a fundamental trade-off between instance detection rates and boundary precision. The finding has important implications for application domains where mask quality matters: SAM3's superior boundary localization may be preferable for robotic manipulation, precise measurement, or downstream geometric analysis, even though YOLO11 models achieve better overall detection performance.}
\label{fig:iou_sensitivity}
\end{figure*}

We next examine how performance varies as a function of the IoU threshold used to match predictions to ground-truth instances. Figure~\ref{fig:iou_sensitivity} presents F1-score curves across ten IoU thresholds from 0.05 to 0.50 for all four models, revealing dramatically different sensitivity patterns between specialized and foundation models.

For YOLO11l, F1-score decreases from 76.4\% at IoU\,=\,0.10 to 71.9\% at 0.15, dropping sharply to 53.4\% at 0.30 and continuing to 26.3\% at 0.50. This corresponds to a 23.0-point drop between IoU\,=\,0.10 and 0.30, and a 50.1-point drop between 0.10 and 0.50, reflecting strong sensitivity of instance-level metrics to threshold choice in dense scenes. YOLO11m exhibits similar behavior, declining from 76.6\% at IoU\,=\,0.10 to 72.2\% at 0.15, 53.9\% at 0.30, and 26.2\% at 0.50 (50.4-point total drop). Even the smallest YOLO11n variant shows this pattern: 73.1\% at 0.10, 68.9\% at 0.15, 52.0\% at 0.30, and 24.8\% at 0.50 (48.3-point drop).

In striking contrast, SAM3 exhibits remarkable stability across IoU thresholds. Starting at 63.8\% F1 at IoU\,=\,0.10, performance remains essentially unchanged at 63.8\% at 0.15, declines modestly to 63.4\% at 0.30, and reaches 59.8\% at 0.50 a total drop of only 4.0 percentage points across the entire range. This 12-fold smaller degradation compared to YOLO11 models suggests that despite lower absolute F1 scores at lenient thresholds, SAM3 produces masks with substantially higher boundary quality and spatial precision (Table~\ref{tab:mask_quality_comparison}).

This finding reveals a fundamental trade-off: YOLO11 models achieve higher instance-level detection rates at relaxed matching criteria but produce masks with less precise boundaries, leading to steep performance degradation as evaluation becomes stricter. SAM3's foundation model pre-training on over one billion masks appears to confer superior boundary localization, maintaining performance even under stringent IoU requirements. For applications where mask quality and boundary precision are critical such as robotic grasping, precise yield estimation, or downstream shape analysis SAM3's stability may outweigh its lower absolute F1 at IoU\,=\,0.15.

\begin{table*}[t]
\centering
\caption{Comparison of YOLO11-seg variants and SAM3 for instance mask quality and IoU sensitivity. Both models output pixel-level instance masks; the primary distinction is boundary precision and the architectural/training causes that drive IoU sensitivity. $\Delta\mathrm{F1}$ reports the change in F1 between IoU=0.10 and IoU=0.50 (positive = drop).}
\label{tab:mask_quality_comparison}
\begin{tabular}{p{2.6cm} p{3.0cm} p{4.2cm} p{4.0cm} p{2.2cm}}
\hline
\textbf{Model Type} & \textbf{Output (format)} & \textbf{Boundary Precision \& Behavior} & \textbf{Mechanistic / Training Reasons} & \textbf{IoU Sensitivity} \\
\hline

\textbf{YOLO11-seg (n/m/l)} &
Pixel-level instance masks &
Masks exhibit coarser boundaries with small but systematic boundary errors (jagged edges, slight over-/under-segmentation, and smoothing/quantization at contours). These small boundary discrepancies reduce intersection area under stricter IoU thresholds. &
Architecture is box-driven detection followed by mask refinement; mask heads often rely on lower-resolution feature maps, upsampling, or class-conditional refinement that can blur fine edges. Training data/augmentation or loss emphasis on detection/coverage rather than per-pixel boundary fidelity can exacerbate boundary noise. &
Large drop: $\Delta\mathrm{F1}\approx$ \textbf{48--50 points} (IoU 0.10 → 0.50). \\

\textbf{SAM3} &
Pixel-level instance masks &
Masks show high-fidelity boundaries with tight alignment to object contours and fewer small-scale boundary errors. Overlap remains robust even as IoU threshold increases. &
SAM-style models are trained with extremely large-scale mask supervision and pixel-accurate objectives, and often include explicit boundary-aware design choices (high-resolution decoders, iterative refinement, mask priors) that preserve fine-grained geometry. &
Small drop: $\Delta\mathrm{F1}\approx$ \textbf{4 points} (IoU 0.10 → 0.50). \\

\hline
\end{tabular}
\end{table*}

\subsection{Analysis of Evaluation Protocol}
\label{sec:results_protocol}

The final set of results concerns the experimental investigations that led to the adopted evaluation protocol, particularly the IoU threshold and confidence settings.

\subsubsection{Initial Performance Discrepancy}

After training YOLO11l with an initial augmentation configuration (mosaic\,=\,0.2, copy-paste\,=\,0.3), the training pipeline reported strong validation metrics: mask mAP$_{50}$ of 75.4\% and box mAP$_{50}$ of 86.9\%. However, when the same model was evaluated on the test set using an instance-level F1 metric at IoU\,=\,0.30, performance dropped to 54.0\% F1 (precision\,=\,56.9\%, recall\,=\,51.3\%). The 21.4-point gap between validation mAP$_{50}$ and test F1 suggested a mismatch between training-time and evaluation-time criteria rather than an obvious failure of the underlying model.

\subsubsection{Confidence Threshold Sweep}

To test whether miscalibrated confidence thresholds were responsible, we swept the YOLO11l confidence threshold from 0.15 to 0.40. Across this range, F1 varied only modestly, from 50.2\% at confidence\,=\,0.15 to 54.0\% at confidence\,=\,0.35--0.40. Although 0.35 provided a slightly better precision–recall balance, the overall performance at IoU\,=\,0.30 remained significantly lower than the training-time mAP$_{50}$, indicating that confidence threshold selection alone could not explain the discrepancy. The threshold of 0.35 was nonetheless fixed for subsequent evaluations of all YOLO11 variants.

\subsubsection{Augmentation and Overfitting Check}
\label{sec:Augmentation_and_Overfitting_Check}
Next, we investigated whether overfitting to the training set caused the observed performance gap. YOLO11l was retrained with substantially stronger data augmentation: mosaic was increased from 0.2 to 0.5 (2.5$\times$), copy-paste from 0.3 to 0.5 (1.67$\times$), and additional mixup\,=\,0.3 and rotation of $\pm 15^{\circ}$ were introduced. This configuration improved validation mask mAP$_{50}$ from 75.4\% to 87.2\%, indicating enhanced feature learning and increased apparent training-time performance. However, the test F1 at IoU\,=\,0.30 remained nearly unchanged at 53.5\%, differing by only 0.5 points from the original 54.0\%. These results strongly suggested that the underlying model capacity and regularization were adequate, and that evaluation settings rather than model overfitting were driving the discrepancy.

\subsubsection{Dataset Distribution Analysis}

To rule out distribution shift between training, validation, and test splits, we conducted a detailed analysis of MinneApple dataset statistics. The three splits exhibited similar characteristics: the training images contained on average 42.1 apples per image, the validation set 45.3, and the test set 38.9. The test set was thus slightly less dense than the training and validation sets, making it, if anything, somewhat easier. This analysis excluded dataset imbalance or covariate shift as the primary cause of the observed mismatch in performance metrics.

\subsubsection{Validation Re-Evaluation and IoU Sensitivity}

The key insight emerged when we re-evaluated the validation set using the same instance-level F1 metric at IoU\,=\,0.30 as on the test set, rather than relying on the training-time mAP$_{50}$ values. Under these conditions, the validation F1 was also low (approximately 42 to 43\% on sampled images), closely resembling the test-set behavior. This indicated that the discrepancy was largely due to the difference between training-time mask-level mAP$_{50}$ and evaluation-time instance-level F1 at a stricter IoU threshold.

A subsequent systematic IoU sweep for YOLO11l, from 0.05 to 0.50, showed that at IoU\,=\,0.10 the model achieved F1\,=\,76.4\%, closely matching the original validation mAP$_{50}$ of 75.4\%. As IoU increased, F1 decreased sharply, reaching 71.9\% at IoU\,=\,0.15 and 53.4\% at IoU\,=\,0.30, and continuing down to 26.3\% at IoU\,=\,0.50. Visual inspection of predictions rejected at higher IoU thresholds confirmed that many of these ``misses'' involved semantically correct detections with minor boundary discrepancies of only a few pixels in dense, heavily overlapping scenes. Notably, the same analysis applied to SAM3 revealed dramatically different behavior: its F1 remained nearly constant at 63.8\% (IoU=0.10 and 0.15), 63.4\% (0.30), and 59.8\% (0.50), indicating superior boundary quality despite lower absolute detection rates.

\subsubsection{Summary of Protocol Choice}

Taken together, the confidence threshold sweep, augmentation experiments, distribution analysis, and IoU sensitivity study indicate that the models themselves are learning meaningful representations and that the primary source of the initial discrepancy lies in the evaluation configuration. The final choice of IoU\,=\,0.15 reflects an empirically grounded compromise: it yields instance-level F1 scores that align well with training-time mAP$_{50}$, while accommodating unavoidable boundary ambiguities in dense MinneApple scenes and still penalizing clearly incorrect detections. All results reported earlier in this section are therefore based on IoU\,=\,0.15 and confidence\,=\,0.35 for YOLO11, and the corresponding tuned settings for SAM3.

\section{Discussion}
\label{sec:discussion}
\begin{figure*}[!htbp]
\centering
\includegraphics[width=\textwidth]{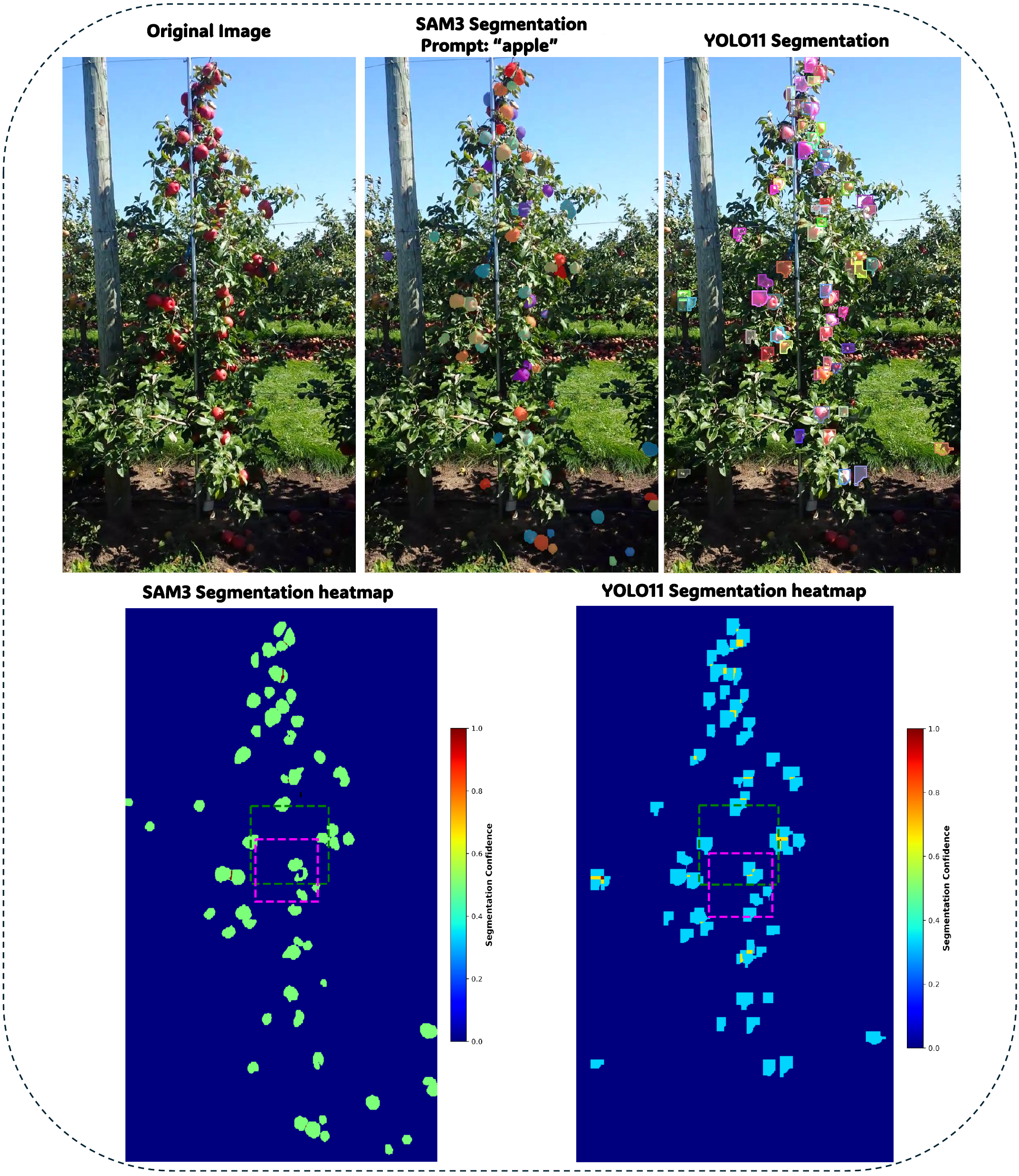}
\caption{Zero-shot SAM3 and fine-tuned YOLO11-seg results in segmenting Ripe apples in Mineapple Dataset. This comparison illustrates specialization-driven recall in YOLO11 versus SAM3’s concept-level boundary precision.}
\label{fig:discussiona}
\end{figure*}

\begin{figure*}[!htbp]
\centering
\includegraphics[width=\textwidth]{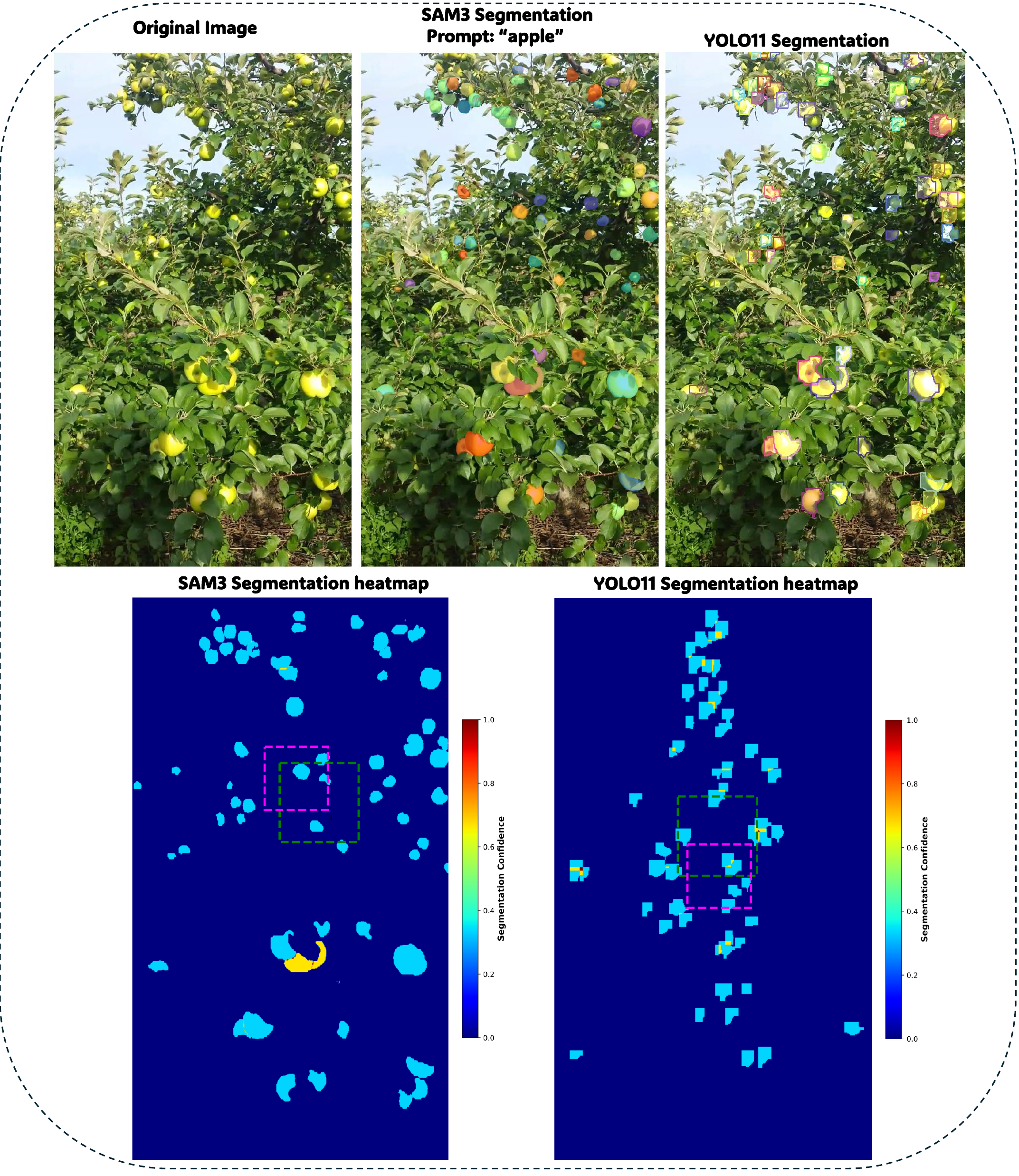}
\caption{Zero-shot SAM3 and fine-tuned YOLO11-seg results in segmenting green apples from Mineapple dataset. Comparison illustrates specialization-driven recall in YOLO11 versus SAM3’s concept-level boundary precision.}
\label{fig:discussionb}
\end{figure*}

The qualitative comparison in Fig.~\ref{fig:discussiona}a and Fig.~\ref{fig:discussionb} further reinforces the quantitative trends observed in our evaluation. In densely packed orchard scenes, fine-tuned YOLO11 exhibits strong specialization by detecting a large number of apple instances, including small or partially occluded fruit; however, its masks often display irregular boundaries and duplicate detections, consistent with its steep IoU-sensitivity profile. In contrast, SAM3 demonstrates more conservative instance counts but produces smoother, semantically coherent masks that adhere closely to object contours, as visible in both the ripe-fruit example (\ref{fig:discussiona}) and the green-fruit scenario (\ref{fig:discussionb}))

These results highlight a key trade-off between foundation-model generalization and detector specialization: YOLO11 excels in recall-driven enumeration, whereas SAM3 maintains geometric precision despite missing instances. Such behaviors underscore the importance of dataset characteristics and task objectives when selecting between zero-shot models and fine-tuned detectors for high-density agricultural segmentation.
\subsection{Specialization Wins for Performance-Critical Applications}
\label{sec:discussion_specialization}

\begin{figure*}[!htbp]
\centering
\includegraphics[width=\textwidth]{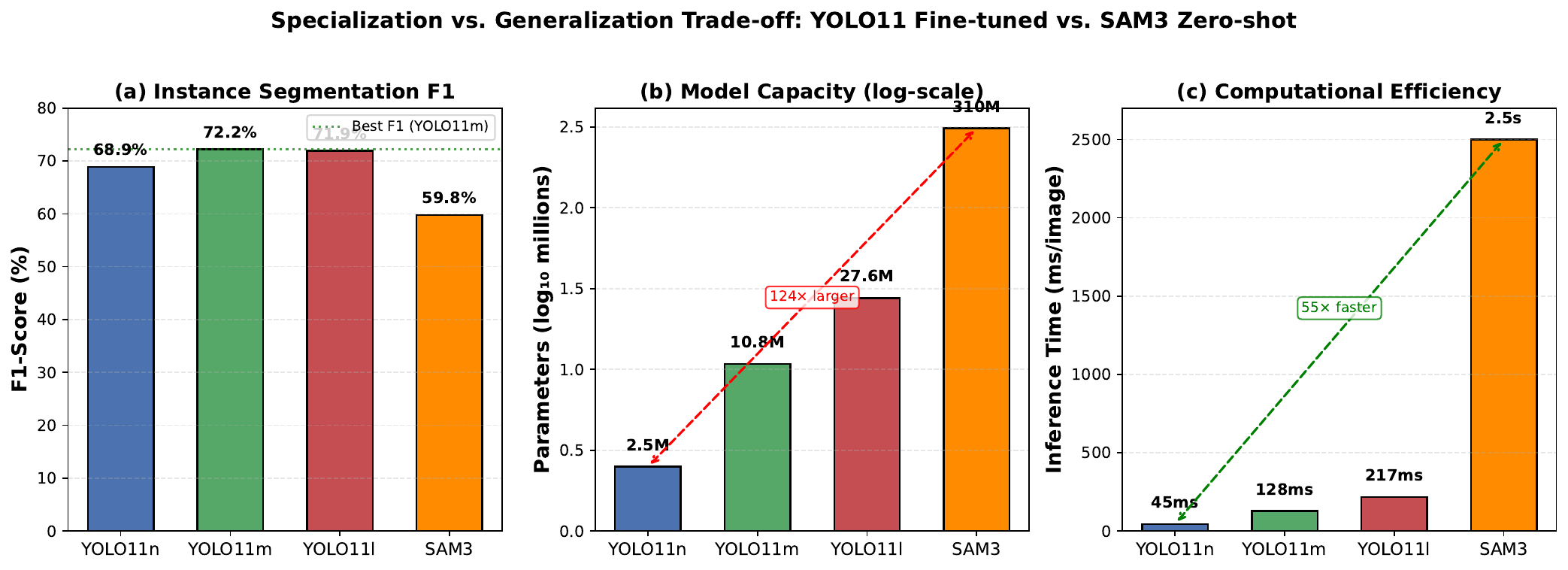}
\caption{Specialization vs. generalization trade-off comparing fine-tuned YOLO11 models and zero-shot SAM3 across three dimensions: (a) instance segmentation F1-score showing YOLO11's performance advantage with YOLO11m achieving the highest F1 at 72.2\%, representing a 12.4 percentage point improvement over SAM3's 59.8\%; (b) model capacity (log$_{10}$-scaled parameters) revealing that SAM3 uses 124$\times$ more parameters than YOLO11n (310M vs. 2.5M) yet underperforms all fine-tuned variants; (c) computational efficiency (inference time per image) demonstrating that YOLO11n processes images 55$\times$ faster than SAM3 (45ms vs. 2500ms). The figure illustrates that task-specific fine-tuning yields simultaneous improvements in both accuracy and computational efficiency: even the smallest YOLO11n model with 2.5M parameters exceeds SAM3 by 9.1 F1 points while requiring dramatically less computation. This counterintuitive result smaller models outperforming massive foundation models highlights the decisive advantage of specialization for performance-critical instance segmentation when domain-specific training data is available.}
\label{fig:bar_f1_params_runtime}
\end{figure*}

The quantitative results clearly indicate that task-specific fine-tuning provides a decisive advantage over zero-shot foundation models for performance-critical instance segmentation (Figure~\ref{fig:bar_f1_params_runtime}). YOLO11m attains an F1-score of 72.2\%, corresponding to a 12.4 percentage point absolute improvement over SAM3 (59.8\%), while using approximately 28.7$\times$ fewer parameters and achieving about 19.5$\times$ lower inference latency. Even the smallest YOLO11n variant, with only 2.5M parameters, reaches 68.9\% F1 and thus still exceeds SAM3 by 9.1 points despite its dramatically smaller capacity.

These results are consistent with broader computer vision trends, where specialized, fine-tuned models typically outperform generalist foundation models on well-defined downstream tasks. Fine-tuning allows the model to adapt to task-specific visual characteristics (e.g., object appearance, occlusion patterns, illumination conditions) and to optimize directly for the target data distribution and evaluation metric. Moreover, specialization concentrates computational capacity on the relevant tasks rather than maintaining broad generalization overhead.

For application domains in which detection accuracy directly influences operational or economic outcomes, such as yield estimation, harvest planning, or disease detection, this performance gap is practically significant. A 12-point F1 improvement translates into hundreds of additional correctly segmented instances and fewer false alarms over a large dataset, which in turn improves downstream estimates and decisions.

\subsection{The Persistent Value of Foundation Models}
\label{sec:discussion_foundation}

Although SAM3 underperforms fine-tuned YOLO11 models under the evaluated conditions, its zero-shot performance remains a notable milestone for generalist computer vision. Achieving 59.8\% F1 on dense instance segmentation without any labeled data, annotation effort, or task-specific optimization would have been competitive with many task-specific architectures only a few years ago. That this level of performance is obtained from a single text prompt (``apple'') underscores the strength of the underlying foundation model and its pre-training on more than one billion masks.

In scenarios where labeled data are scarce, costly, or slow to obtain e.g., rare crops, emerging diseases, or rapidly changing field conditions, this zero-shot capability offers immediate utility. SAM3’s 310M parameters encode a strong prior over object shapes, textures, and boundaries across diverse visual domains, which can transfer to specialized tasks it has never explicitly seen during training. In addition, its architecture supports interactive prompts (points, boxes, and text), enabling human-in-the-loop correction for specific instances where high precision is required at the local level rather than globally.

\begin{figure*}[!htbp]
\centering
\includegraphics[width=\textwidth]{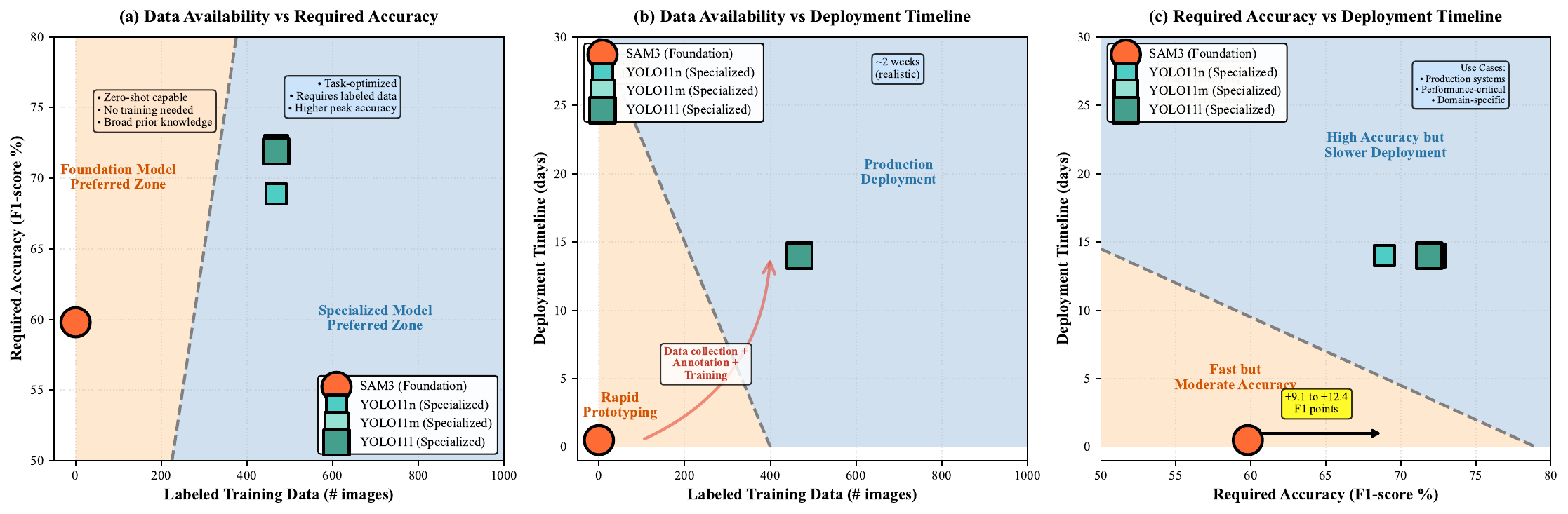}
\caption{Complementary roles of foundation models (SAM3) and specialized models (YOLO11) across three deployment dimensions. Panel (a) shows data availability versus required accuracy: SAM3 achieves 59.8\% F1 with zero training data, while YOLO11 variants require 468 labeled images but deliver 68.9--72.2\% F1. Panel (b) illustrates deployment timeline: SAM3 enables immediate inference ($<$1 day) versus $\sim$14 days for specialized model deployment. The $\sim$14 days comprises: data collection ($\sim$7 days), semi-automated annotation of 28,179 instances across 468 images ($\sim$5 days), infrastructure setup and validation ($\sim$2 days), and training itself ($<$15 minutes measured). Panel (c) directly compares accuracy against time-to-deployment, showing foundation models provide moderate accuracy instantly while specialized models achieve 9--12 percentage points higher F1 with substantial upfront investment. Shaded regions and decision boundaries (gray dashed) indicate where each model family is most appropriate given project constraints on data, time, and accuracy requirements.}
\label{fig:foundation_vs_specialized_roles}
\end{figure*}

From a system design perspective, foundation models are also highly valuable for rapid prototyping and exploratory analysis: they can provide baseline segmentation performance and qualitative insights before substantial data collection and training infrastructure are in place. Furthermore, their broad pre-training may confer robustness to distribution shifts (e.g., new varieties, seasons, or lighting conditions) that could degrade the performance of narrowly trained specialized models. An important open question is whether SAM3’s zero-shot performance generalizes more gracefully across different orchards, cultivars, or imaging protocols than YOLO11 models fine-tuned on a single dataset. These complementary roles are visualized in Figure~\ref{fig:foundation_vs_specialized_roles}.

\subsection{The Critical Importance of IoU Threshold Selection}
\label{sec:discussion_iou}

\begin{figure*}[!htbp]
\centering
\includegraphics[width=0.9\textwidth]{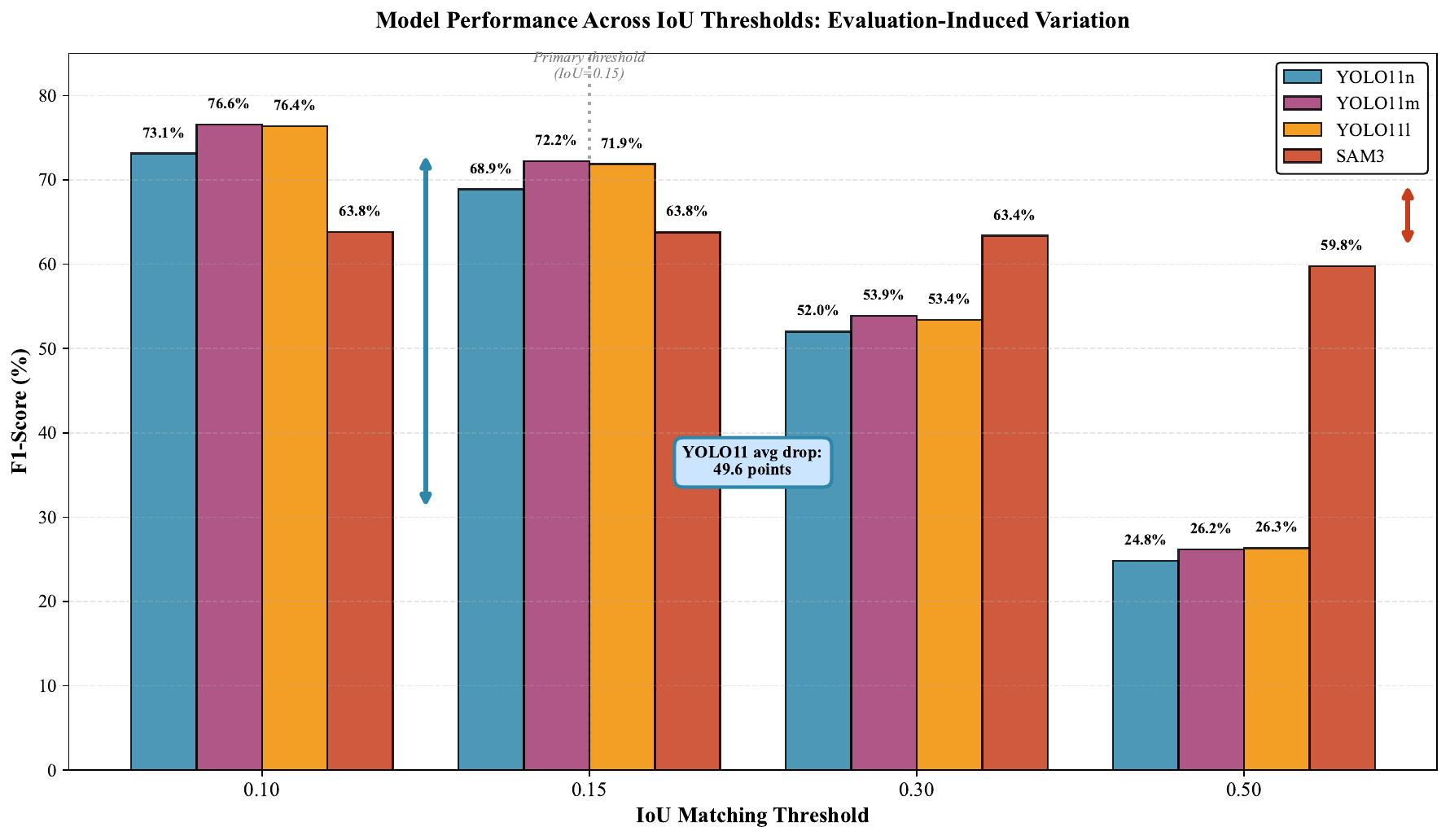}
\caption{Side-by-side comparison of F1-scores at four IoU matching thresholds (0.10, 0.15, 0.30, 0.50) for all models, illustrating the magnitude of evaluation-induced performance variation. YOLO11 variants exhibit steep degradation as IoU threshold increases: YOLO11n drops 48.3 points (73.1\% to 24.8\%), YOLO11m drops 50.4 points (76.6\% to 26.2\%), and YOLO11l drops 50.1 points (76.4\% to 26.3\%) between IoU=0.10 and 0.50. In striking contrast, SAM3 shows only a 4.0-point drop (63.8\% to 59.8\%) across the same range, demonstrating 12-fold lower sensitivity. The vertical dashed line marks IoU=0.15, the primary evaluation threshold adopted for dense MinneApple scenes. This visualization makes immediately apparent that small changes in evaluation criteria can induce larger performance differences than model architecture choice itself, and reveals the fundamental trade-off between instance detection rates (YOLO11's strength) and boundary precision (SAM3's strength).}
\label{fig:iou_bar_models}
\end{figure*}

One of the major findings from this study is that IoU threshold selection can have a more dramatic impact on reported performance than the choice of model architecture itself, especially in dense segmentation tasks (Figure~\ref{fig:iou_bar_models}). For YOLO11l, F1-score declines from 76.4\% at IoU\,=\,0.10 to 53.4\% at IoU\,=\,0.30, a drop of 23.0 points. Extending to IoU\,=\,0.50 yields 26.3\% F1, representing a 50.1-point collapse relative to IoU\,=\,0.10. These large swings arise even though the underlying predictions remain visually reasonable for many instances; small boundary misalignments in densely packed objects are heavily penalized at high IoU thresholds.

Remarkably, SAM3 exhibits fundamentally different behavior: its F1-score drops by only 4.0 points from 63.8\% at IoU\,=\,0.10 to 59.8\% at IoU\,=\,0.50, demonstrating 12-fold lower threshold sensitivity than YOLO11 models. This stability reveals that SAM3's foundation model pre-training produces masks with substantially higher boundary precision, even though its absolute instance detection rate is lower. The finding suggests that comparing models solely at a single IoU threshold (such as 0.15 or 0.50) obscures important qualitative differences: YOLO11 excels at instance-level detection but produces spatially imprecise masks, whereas SAM3 generates fewer detections but with superior boundary localization.

This has direct implications for benchmarking. Standard COCO-style thresholds (e.g., IoU\,=\,0.50) are appropriate when objects are large, isolated, and cleanly separated, but they are overly strict for densely packed, overlapping objects with inherently ambiguous boundaries. For MinneApple-style dense scenes, our analysis suggests that IoU\,=\,0.15 provides a more balanced operating point for comparing instance detection capabilities: it maintains a requirement for substantial overlap between prediction and ground-truth while accommodating boundary variability due to occlusion and annotation uncertainty. However, the stark difference in IoU sensitivity between model families indicates that multi-threshold evaluation is essential for understanding true model behavior.

To understand the practical implications of IoU\,=\,0.15, consider its spatial tolerance in terms of centroid localization error. For a typical MinneApple apple with diameter $d \approx 60$ pixels (radius $r = 30$ pixels), two circular masks achieve IoU\,=\,0.15 when their centers are displaced by approximately $1.3r$ to $1.4r$, corresponding to a maximum center-to-center error of $\sim$40--42 pixels, or roughly 67--70\% of the object's diameter. This substantial positional tolerance explains why IoU\,=\,0.15 is appropriate for detection and counting tasks: a prediction is considered correct as long as it identifies the presence and approximate location of an apple, even if the mask boundary is imprecise. However, this same tolerance is insufficient for applications requiring accurate spatial localization. For instance, robotic grasping systems typically require centroid accuracy within $\pm$10--15 pixels to ensure successful grasp execution; at IoU\,=\,0.15, centroid errors can be 2.5--4$\times$ larger than this requirement, making the threshold inadequate for manipulation tasks. Similarly, geometric measurements such as fruit volume or weight estimation depend on accurate boundary delineation and would incur 15--25\% errors under the lenient boundary constraints of IoU\,=\,0.15.

More broadly, evaluation protocols should be explicitly tailored to downstream task requirements. For yield estimation problems, the primary goal is to count instances; modest boundary errors have little practical effect, and lower IoU ranges (0.10 to 0.20) are likely more appropriate. For robotic manipulation or grasp planning, where precise object shape and contact points matter, stricter IoU thresholds (0.30--0.40) may be justified, even at the cost of lower aggregate F1. Reporting performance across multiple thresholds and summarizing results as curves or bands, rather than single scalars, provides a more complete view of model behavior.

\subsubsection{Boundary Quality versus Detection Accuracy}

The IoU sensitivity analysis reveals a previously unrecognized trade-off between instance detection rates and boundary quality. YOLO11 models optimize for high recall and precision at the instance level, successfully identifying most apple instances in dense scenes but producing masks with less precise boundaries. This leads to performance that is highly dependent on evaluation criteria: excellent results at lenient thresholds but steep degradation as boundary precision requirements increase.

In contrast, SAM3's foundation model architecture, pre-trained on over one billion diverse masks, appears to prioritize boundary localization accuracy over detection completeness. Its high stability across IoU thresholds only a 4.0 percentage point drop (6.3\% relative degradation) compared to 48--50 points (65--68\% relative degradation) for YOLO11 suggests that when SAM3 detects an instance, the resulting mask boundary closely matches ground truth. This characteristic may stem from SAM3's training objective and architecture, which emphasize pixel-accurate segmentation rather than bounding-box-driven detection followed by mask refinement as in YOLO-style models.

It is worth noting that two-stage architectures such as Mask R-CNN~\citep{he2017mask} and its successors (Cascade Mask R-CNN~\citep{cai2018cascade}, Hybrid Task Cascade~\citep{chen2019hybrid}) may occupy an intermediate position in this detection-boundary trade-off. These models employ dedicated high-resolution mask branches with RoIAlign for spatial precision~\citep{he2017mask}, and have been shown to produce higher-quality boundaries than single-stage detectors~\citep{bolya2019yolact,wang2020solov2}. Prior work on dense fruit detection suggests that Mask R-CNN variants can achieve 5--10 percentage point better IoU stability than YOLO-style models~\citep{koirala2019deep,bargoti2017deep}, though still falling short of SAM-level boundary fidelity due to differences in training data scale (thousands vs.\ billions of masks) and architectural emphasis on iterative refinement~\citep{kirillov2023segment}. Future comparative studies including two-stage detectors would help clarify the full spectrum of detection-boundary trade-offs in agricultural instance segmentation.

For practical deployment, this trade-off has important implications. Applications requiring accurate instance counts such as yield estimation, inventory management, or population monitoring benefit from YOLO11's higher detection rates at IoU=0.15, despite boundary imprecision. Conversely, applications requiring precise spatial information such as robotic harvesting (grasp point localization), geometric measurement (fruit size/shape), or dense 3D reconstruction may benefit from SAM3's superior boundary quality, even if some instances are missed. The choice between specialized and foundation models should therefore consider not only aggregate F1 scores but also the relative importance of detection completeness versus boundary fidelity for the target application.

The path to this insight also highlights the importance of aligning training and evaluation metrics. Initially, YOLO11’s strong validation mAP$_{50}$ (75-87\%) appeared inconsistent with test F1 around 54\% at IoU\,=\,0.30, raising concerns about overfitting. Only after re-evaluating the validation set with the same instance-level metric and threshold as the test set, and conducting a systematic IoU sweep, did the discrepancy resolve as an evaluation mismatch rather than a modeling failure.

\subsection{Scaling Laws and Model Capacity Trade-offs}
\label{sec:discussion_scaling}

\begin{figure*}[!htbp]
\centering
\includegraphics[width=0.99\textwidth]{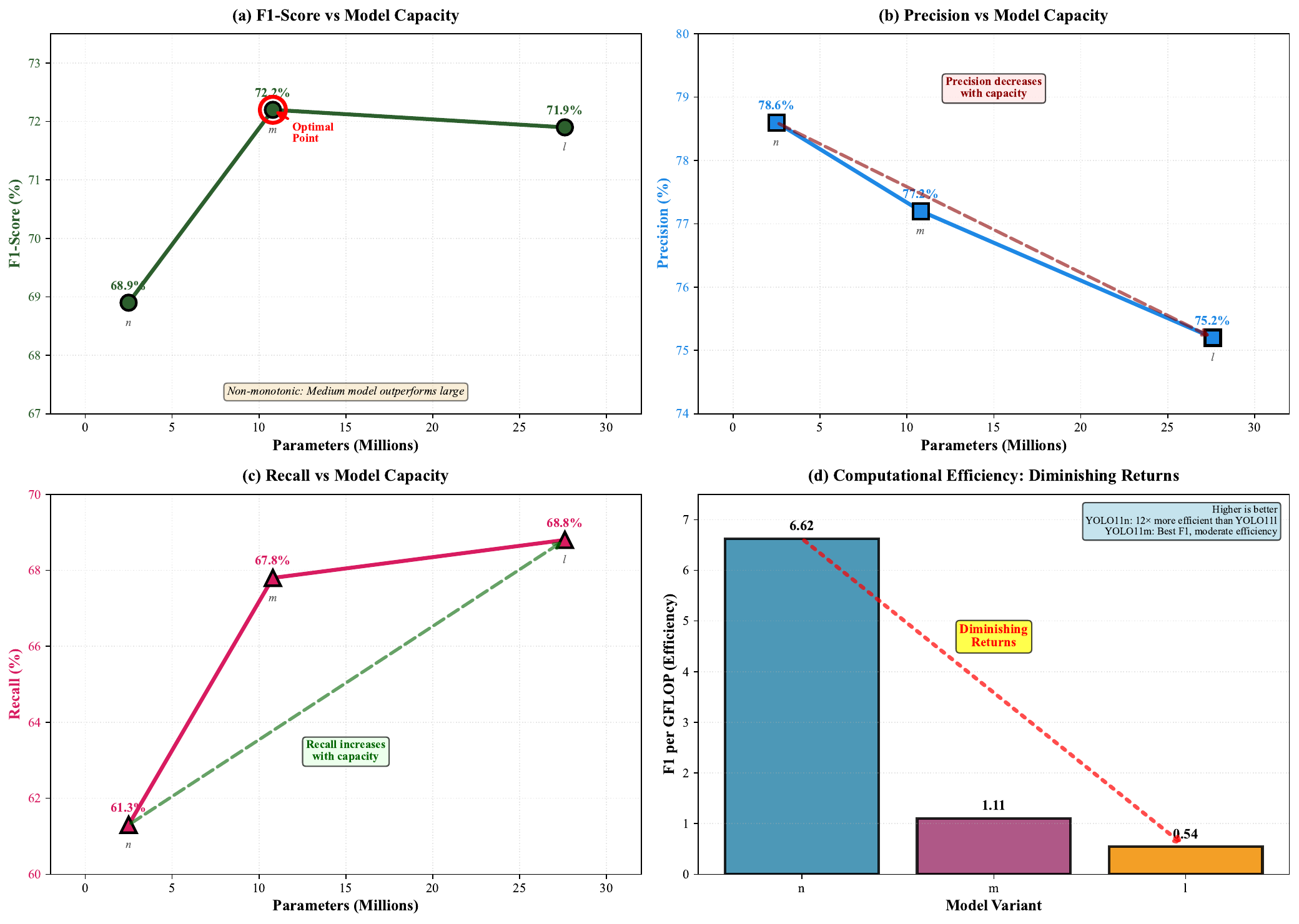}
\caption{Scaling behavior of YOLO11 variants showing non-monotonic relationship between model capacity and performance on MinneApple's limited training data (468 images). Panel (a) shows F1-score versus parameter count, revealing that YOLO11m (10.8M parameters, 72.2\% F1) achieves peak performance, slightly outperforming the larger YOLO11l (27.6M parameters, 71.9\% F1) despite having 2.5$\times$ fewer parameters. Panel (b) demonstrates precision degradation with increasing capacity (78.6\% $\rightarrow$ 77.2\% $\rightarrow$ 75.2\%), while panel (c) shows recall improvement (61.3\% $\rightarrow$ 67.8\% $\rightarrow$ 68.8\%), indicating that larger models detect more difficult instances at the cost of additional false positives. Panel (d) quantifies computational efficiency as F1 per GFLOP, revealing dramatic diminishing returns: YOLO11n achieves 6.62 F1/GFLOP compared to YOLO11l's 0.54 F1/GFLOP, representing a 12$\times$ efficiency advantage for only a 3-point F1 sacrifice. The non-monotonic F1 curve and efficiency degradation suggest that medium-capacity architectures offer optimal bias--variance trade-offs in limited-data regimes, with models exceeding $\sim$11M parameters insufficiently constrained by the available training data.}
\label{fig:scaling_law_plot}
\end{figure*}

The comparison among YOLO11n, YOLO11m, and YOLO11l reveals a nuanced scaling pattern (Figure~\ref{fig:scaling_law_plot}). YOLO11l has approximately 11$\times$ more parameters than YOLO11n (27.6M vs.\ 2.5M), yet delivers only a 3.0-point F1 improvement (71.9\% vs.\ 68.9\%). YOLO11m, with intermediate capacity (10.8M parameters), achieves the highest F1 of 72.2\%, slightly outperforming YOLO11l. This non-monotonic relationship between model size and F1 suggests that, given the MinneApple training data (468 images), the medium-capacity model may be closest to an optimal bias–variance trade-off, whereas the largest variant is not fully utilized.

Precision–recall profiles support this interpretation. YOLO11l reaches the highest recall (68.8\%) but the lowest precision (75.2\%), indicating that increased capacity primarily boosts the ability to detect more difficult instances, but at the cost of more false positives. YOLO11n exhibits the opposite behavior, with the highest precision (78.6\%) and lowest recall (61.3\%), reflecting a more conservative decision boundary. YOLO11m sits between these extremes (77.2\% precision, 67.8\% recall), delivering the most balanced performance and thus the highest F1.

These observations are consistent with scaling laws in limited-data regimes, confirming past findings in object detection~\citep{tan2020efficientdet,zhu2021deformable} and agricultural vision~\citep{koirala2019deep,tian2019apple}: after a certain point, adding parameters yields diminishing returns in accuracy and may even degrade calibration or robustness if data are insufficient to constrain the model. For datasets of the order of a few hundred images, medium-sized architectures may therefore offer better accuracy–efficiency trade-offs than very large models. Future work should investigate whether YOLO11l’s recall advantage could be further exploited with additional training data or tailored regularization strategies.

\begin{figure*}[ht!]
\centering
\includegraphics[width=0.98\textwidth]{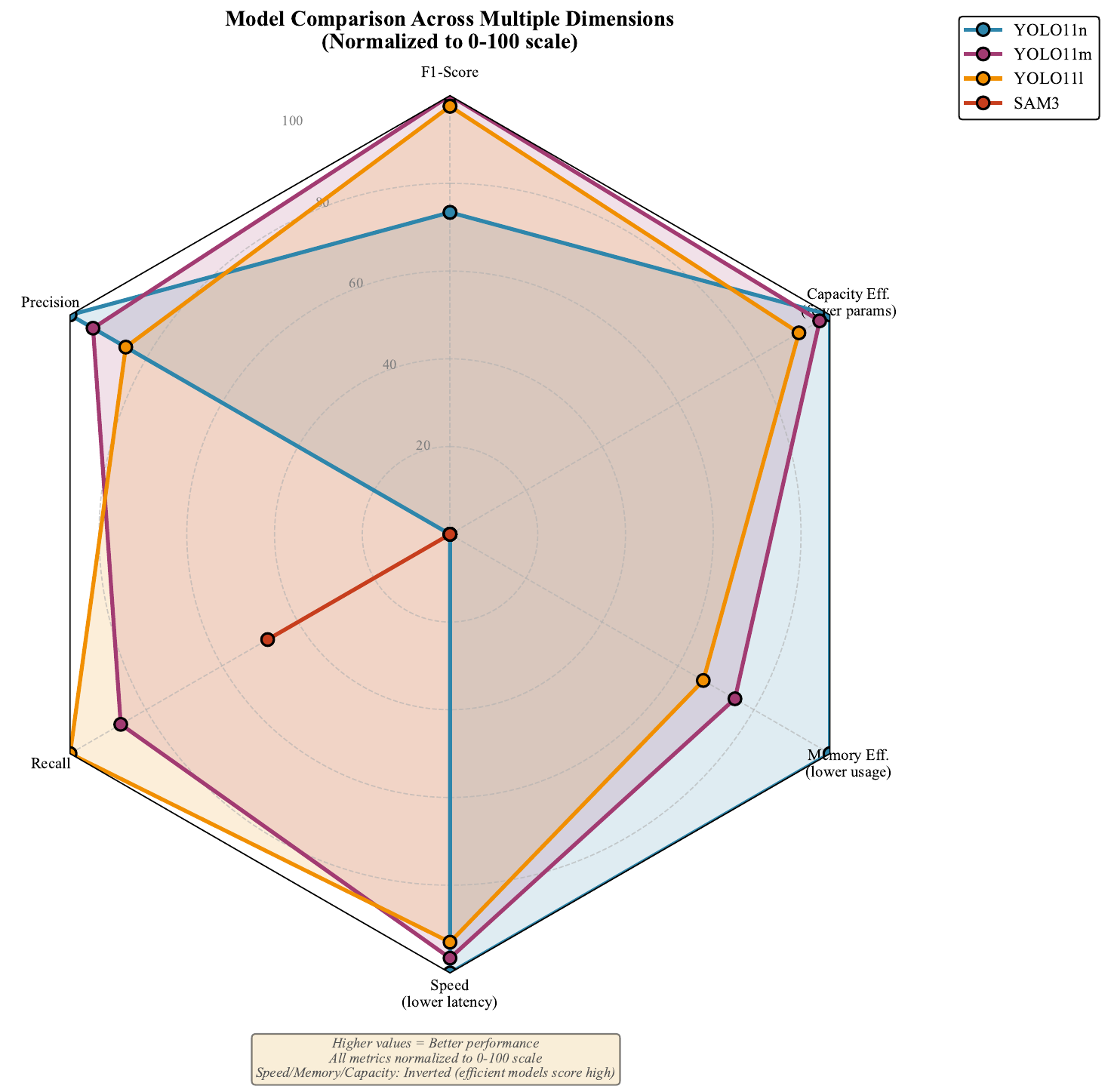}
\caption{Radar plot comparing SAM3 and YOLO11 variants across six normalized metrics (0–100): F1-score, precision, recall, speed (inverse latency), memory efficiency (inverse usage), and capacity efficiency (inverse parameters). YOLO11n (blue) shows maximal efficiency (speed = 100, memory = 100, capacity = 100) with moderate F1 (68.9\%). YOLO11m (purple) provides the best balance with peak F1 (72.2\%). YOLO11l (orange) offers highest recall (68.8\%) but lower efficiency. SAM3 (red) shows strong recall (64.9\%) but minimal efficiency (speed = 2, memory = 50, capacity = 1) due to 310M parameters and 2,500 ms inference.}
\label{fig:radar_tradeoffs}
\end{figure*}
\subsection{Practical Recommendations for Deployment}
\label{sec:discussion_practical}

The empirical findings naturally translate into deployment guidelines for practitioners designing real-world vision systems (Figure~\ref{fig:radar_tradeoffs} provides a visual decision aid):

\begin{itemize}
    \item \textbf{Production deployments with labeled data available.} Fine-tuned YOLO11 models should be the primary choice when accuracy is critical and domain-specific training data can be collected. YOLO11m, in particular, provides the best accuracy–efficiency compromise for datasets of approximately 400-600 images: its 72.2\% F1, 128\,ms inference time (about 7.8 FPS), and moderate memory footprint permit real-time operation on many edge devices. In our experiments, however, we used more computational resources than were strictly necessary (see Section~\ref{sec:yolo_infra}).
    \item \textbf{Resource-constrained or ultra-low-latency scenarios.} YOLO11n offers a favorable operating point when computational budget is tight. With 2.5M parameters, 45\,ms inference (about 22 FPS), and 68.9\% F1, it is suitable for embedded systems, mobile platforms, or high-throughput settings where some loss in F1 relative to YOLO11m is acceptable.
    \item \textbf{Rapid prototyping and zero-shot use cases.} SAM3 is attractive when labeled data, training infrastructure, or specialized ML expertise are limited. Its 59.8\% F1 in zero-shot mode is adequate for exploratory studies, preliminary benchmarking, or applications with lower accuracy requirements. Its interactive prompting interface also supports workflows where human operators iteratively refine segmentations for specific frames or regions.
    \item \textbf{Evaluation methodology best practices.} IoU thresholds should be set in line with downstream task needs and dataset characteristics (e.g., IoU\,=\,0.15 for dense MinneApple-style scenes), and performance should be reported over a range of thresholds rather than a single value. Crucially, training-time metrics and test-time metrics must be aligned to avoid misleading interpretation of validation–test discrepancies.
\end{itemize}

\begin{figure*}[ht!]
\centering
\includegraphics[width=0.99\textwidth]{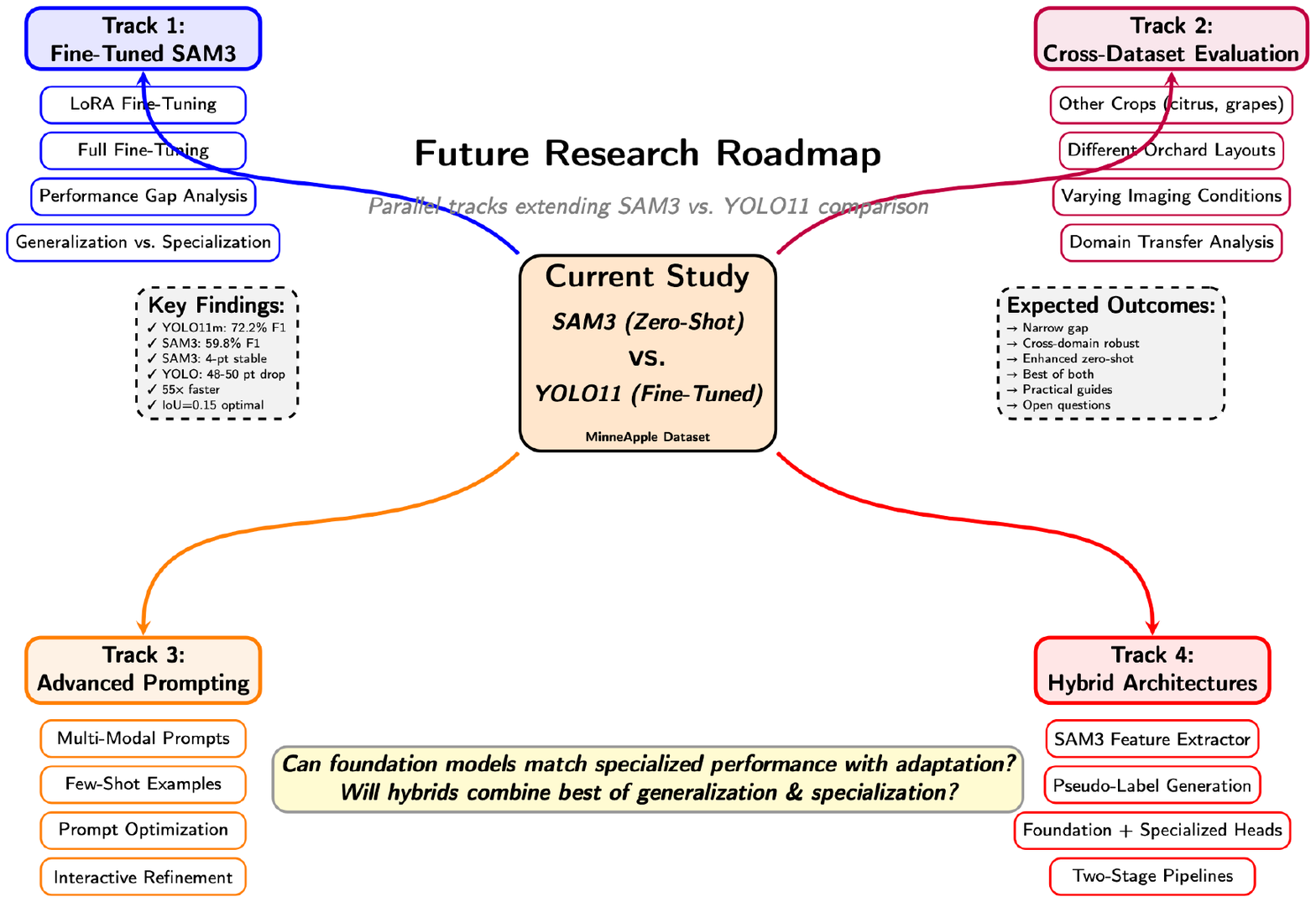}
\caption{Future research directions roadmap showing four parallel tracks extending the current SAM3 vs. YOLO11 comparison study. Track 1 (blue) explores fine-tuned SAM3 variants including LoRA and full fine-tuning to narrow the 12-point F1 gap. Track 2 (purple) investigates cross-dataset generalization to other crops, orchard layouts, and imaging conditions. Track 3 (orange) develops advanced prompting strategies combining multi-modal inputs and few-shot learning. Track 4 (red) proposes hybrid SAM-YOLO architectures that combine foundation model priors with specialized detector efficiency. The central node summarizes current findings (YOLO11m: 72.2\% F1, SAM3: 59.8\% zero-shot, 55$\times$ speed advantage), while supporting boxes highlight key results and expected outcomes. Open questions address whether foundation models can match specialized performance through adaptation and whether hybrids can optimally combine generalization and specialization capabilities.}
\label{fig:future_work_diagram}
\end{figure*}

\subsection{Limitations and Future Work}
\label{sec:discussion_limitations}

Several limitations of this study suggest promising avenues for future research. First, we considered only SAM3 in a pure zero-shot setting. Parameter-efficient fine-tuning techniques such as LoRA, as well as full fine-tuning, could significantly narrow or even eliminate the performance gap relative to YOLO11 while preserving some of SAM3’s generalization and interactive capabilities. A systematic comparison between LoRA-tuned SAM3 and fully fine-tuned YOLO11 models would clarify the trade-offs between adapting foundation models and training specialized detectors from pre-trained weights.

Additionally, the evaluation focused on a single dataset (MinneApple) and one object class (apples). It remains to be seen whether the observed specialization - generalization trade-offs and IoU sensitivity patterns generalize to other crops, orchard layouts, imaging modalities, or environmental conditions. Cross-dataset and cross-domain experiments would be needed to test robustness and transferability.

Moreover, we restricted our prompting strategy for SAM3 to a single, simple text prompt (``apple''). More elaborate prompting, potentially combining text, example masks, and bounding boxes, could further enhance zero-shot performance and may close part of the gap to fine-tuned models. Exploring multi-modal prompting strategies and prompt optimization is therefore, an important direction.

Furthermore, hybrid architectures that combine foundation models’ semantic priors with the efficiency and accuracy of specialized detectors are a natural next step. Possible approaches include using SAM3 as a feature extractor feeding a lighter detection head, or employing SAM3 to generate pseudo-labels or proposals that bootstrap YOLO-style training. Such hybrids could exploit both broad generalization and task-specific specialization within a unified system. Figure~\ref{fig:future_work_diagram} outlines these and other promising research directions.

\section{Conclusion}

This study presented a systematic comparison between the zero-shot SAM3 foundation model and fine-tuned YOLO11 variants for dense apple instance segmentation on the MinneApple dataset. The results reveal a clear specialization generalization trade-off that informs the design of future agricultural vision systems. Despite receiving no task-specific training, SAM3 achieved a notable 59.8\% F1, demonstrating that modern foundation models now provide practical utility even in specialized agricultural applications.

Fine-tuned YOLO11 models, however, delivered 9--12 percentage point improvements (68.9 to 72.2\% F1), confirming that specialization remains essential when accuracy and efficiency are critical. The smallest 2.5M-parameter YOLO11n still outperformed the 310M-parameter SAM3, underscoring the effectiveness of task-optimized architectures. Among variants, YOLO11m offered the best balance of accuracy and computational cost, achieving 72.2\% F1 with real-time 128\,ms inference and strong precision--recall performance.

A central methodological finding is the decisive influence of IoU threshold selection in dense orchard scenes. Standard thresholds (e.g., IoU=0.50) proved overly strict, masking true model behavior due to boundary ambiguity and occlusion. Through systematic threshold sensitivity analysis, we identified IoU=0.15 as a more appropriate and domain-aligned criterion. This analysis also revealed a critical qualitative difference: SAM3 exhibited exceptional boundary stability, with only a 4-point F1 drop from IoU = 0.10 to 0.50, whereas YOLO11 variants degraded by 48 to 50 points, indicating superior mask precision in the foundation model despite lower detection completeness.

Overall, the complementary strengths of YOLO11 and SAM3 suggest promising hybrid directions, including parameter-efficient tuning and architectures that combine foundation-model priors with specialized detection heads. This work provides practical guidance for selecting and evaluating segmentation models in orchards and similar dense environments, while highlighting methodological considerations essential for robust agricultural computer vision.

\section*{Acknowledgement} This work was supported in part by the National Science Foundation (NSF); in part by United States Department of Agriculture (USDA); in part by the National Institute of Food and Agriculture (NIFA), through the “Artificial Intelligence (AI) Institute for Agriculture” Program, Accession Num ber 1029004 for the Project Titled “Robotic Blossom Thinning with Soft Manipulators” under Award AWD003473, Award AWD004595, and Award USDA-NIFA; and in part by United States Department of Agriculture National Science Foundation (USDANSF), Accession Number 1031712, under the Project “ExPanding University of Central Florida (UCF) AI Research To Novel Agricultural EngineeRing Applications (PARTNER)” under Grant 2024-67022-41788.  Additionally, this work was supported in part by the Intramural Research Program of the U.S. Department of Agriculture (USDA), National Institute of Food and Agriculture, under Grant No. 2024-67022-41788. The views and conclusions expressed in this paper are those of the authors and do not necessarily reflect the official policies or positions of the USDA or the U.S. Government.

\section*{Declarations}
The authors declare no conflicts of interest.

\section*{Statement on AI Writing Assistance}
ChatGPT and Grammarly were utilized to enhance grammatical accuracy and refine sentence structure; all AI-generated revisions were thoroughly reviewed and edited for relevance.

\bibliographystyle{cas-model2-names}

\bibliography{references}



\printcredits




\end{document}